\documentclass[journal]{IEEEtran}
\usepackage{amsmath,amsfonts}
\usepackage{algorithmic}
\usepackage{algorithm}
\usepackage{array}
\usepackage[caption=false,font=normalsize,labelfont=sf,textfont=sf]{subfig}
\usepackage{textcomp}
\usepackage{stfloats}
\usepackage{url}
\usepackage{verbatim}
\usepackage{graphicx}
\usepackage{cite}
\usepackage{amsmath,amsfonts}
\usepackage{algorithmic}
\usepackage{algorithm}
\usepackage{array}
\usepackage[caption=false,font=normalsize,labelfont=sf,textfont=sf]{subfig}
\usepackage{textcomp}
\usepackage{stfloats}
\usepackage{url}
\usepackage{verbatim}
\usepackage{graphicx}
\usepackage{cite}

\usepackage{bm}
\usepackage{amssymb}
\usepackage{color}
\usepackage{multirow}
\usepackage{times}
\usepackage{amsthm}
\usepackage{booktabs}
\usepackage{float}

\newtheorem{assumption}{Assumption}

\newtheorem{lemma}{Lemma}

\newtheorem{theorem}{Theorem}
\newtheorem{remark}{Remark}

\begin{document}

\title{Sparse Federated Learning with Hierarchical Personalization Models}

\author{Xiaofeng Liu, Qing Wang, Yunfeng Shao, Yinchuan Li

\thanks{Xiaofeng Liu and Qing Wang are with School of Electrical and Information Engineering, Tianjin University, Tianjin, China (e-mail: xiaofengliull@tju.edu.cn, wangq@tju.edu.cn). This work was completed during Xiaofeng Liu's internship at the Huawei Noah's Ark Lab for Advanced Study.}
\thanks{Yinchuan Li and Yunfeng Shao are with Huawei Noah's Ark Lab, Beijing, China (e-mail: liyinchuan@huawei.com, shaoyunfeng@huawei.com)}
\thanks{Corresponding author: Yinchuan Li.}
}

\maketitle

\begin{abstract}
Federated learning (FL) can achieve privacy-safe and reliable collaborative training without collecting users' private data. Its excellent privacy security potential promotes a wide range of FL applications in Internet-of-Things (IoT), wireless networks, mobile devices, autonomous vehicles, and cloud medical treatment. 
However, the FL method suffers from poor model performance on non-i.i.d. data and excessive traffic volume.
\textcolor{black}{To this end,} we propose a personalized FL algorithm using a hierarchical proximal mapping based on \textcolor{black}{the moreau envelop}, named sparse federated learning with hierarchical personalized models (sFedHP), which significantly improves the global model performance facing diverse data. A \textcolor{black}{continuously differentiable} approximated $\ell_1$-norm is also used as the sparse constraint to reduce the communication cost. Convergence analysis shows that sFedHP's convergence rate is state-of-the-art with linear speedup and the sparse constraint only reduces the convergence rate to a small extent while significantly reducing the communication cost. Experimentally, we demonstrate the benefits of sFedHP compared with the FedAvg, HierFAVG (hierarchical FedAvg), and personalized FL methods based on local customization, including FedAMP, FedProx, Per-FedAvg, pFedMe, and pFedGP.
\end{abstract}


\section{Introduction}\label{sec:introduction}
\IEEEPARstart{M}{achine} learning methods have proliferated in real-life applications thanks to the tremendous number of labeled samples~\cite{lecun2015deep}. Typically, these samples collected on users' devices, such as mobile phones, are expected to send to a centralized server with mighty computing power to train a deep model~\cite{chen2019deep}. However, users are often reluctant to share personal data due to privacy concerns, which motivates the emergence of federated learning (FL)~\cite{mcmahan2017communication}. 
Federated Averaging (FedAvg)~\cite{mcmahan2017communication} is known as the first FL algorithm to build a global model for different clients while protecting their data locally. Moreover, FL has been used in the Internet of Things (IoT), wireless networks, mobile devices, autonomous vehicles, and cloud medical treatment for its excellent potential in privacy security~\cite{9233457, 9709603, 9134426, 9210812, 9477571, 9079513, ferrag2021federated, khan2021federated, wang2021federated, he2021edge}.

Unfortunately, the distribution of local data stored in different clients varies greatly, and FedAvg performs unwise when meeting none independent and identically distributed (non-i.i.d.) data. In particular, generalization errors of the global model increase significantly with the data's statistical diversity increasing~\cite{li2019fedmd,deng2020adaptive}. To address this problem, personalized FL based on multi-task learning~\cite{sattler2021clustered} and personalization layers~\cite{arivazhagan2019federated}, and local customization~\cite{li2018federated, t2020personalized,
huang2021personalized, fallah2020personalized, achituve2021personalized} have been proposed.
{\color{black}
Most personalized federated learning algorithms prioritize the personalized model performance of individual clients, while overlooking the performance of the global model. However, this approach contradicts the fundamental purpose of federated learning, which aims to build a high-quality global model. Disregarding the global model's performance may lead to a suboptimal model and hinder the inclusion of new clients. Additionally, frequent communication between clients and the server is typically necessary in federated learning to ensure convergence performance, which can be hampered by high latency and limited bandwidth.
}

\textcolor{black}{
To address these challenges, we introduce a novel hierarchical personalized federated learning framework. Our approach includes a personalized edge server that minimizes differences between models during global model aggregation, leading to significant improvements in global model performance and allowing for personalized user models. Furthermore, our hierarchical personalized federated learning architecture covers the client-edge-cloud, reducing direct communication between clients and the cloud server and greatly decreasing communication overhead.
}

\subsection{Main Contributions}

{\bf{Our main contributions}} in this paper are summarized as follows:

{\color{black}
(1) We propose a noval personalized FL framework, named sparse federated learning with hierarchical personalization models ({\texttt{sFedHP}}). Our approach employs a hierarchical proximal mapping technique based on the moreau envelop. This method separates the optimization of client and edge models from that of the global model, promoting personalized models for clients and edge servers while keeping them close to the reference model. As a result, our approach enhances the performance of the global model on non-i.i.d data.

(2) The hierarchical architecture of {\texttt{sFedHP}} reduces direct communication between clients and the cloud server, leading to a significant decrease in communication overhead. The continuously differentiable approximated $\ell_1$-norm constraints in {\texttt{sFedHP}} with sparse version generate sparse models, which further reduce communication costs.
}

(3) We present the convergence analysis of {\texttt{sFedHP}} by exploiting the convexity-preserving and smoothness-enabled properties of the loss function, which characterizes two notorious issues (client-sampling and client-drift errors) in FL~\cite{karimireddy2020scaffold}. With carefully tuned hyperparameters, theoretical analysis shows that sFedHP's convergence rate is state-of-the-art with linear speedup.

\textcolor{black}{
(4) We empirically evaluate the performance of {\texttt{sFedHP}} using different datasets that capture the statistical diversity of clients’ data. We show that {\texttt{sFedHP}} obtained the state-of-the-art performance while greatly reducing the number of parameters by 80\%.  Moreover, {\texttt{sFedHP}} with non-sparse version outperforms  FedAvg, the hierarchical FedAvg~\cite{liu2020client}, and other local customization based personalized FL methods~\cite{li2018federated,fallah2020personalized,huang2021personalized,achituve2021personalized} in terms of global model.
}

{\color{black}
\subsection{Organization and Main Notation}
The remainder of the paper is organized as follow. Section~\ref{sec_related_work} undertake a complete literature review to illustrate the existing research findings. The problem formulation and algorithm are formulated in Section~\ref{sec_2}. Section~\ref{sec_3} shows the convergence analysis with some important Lemmas and Theorems. Section~\ref{sec_results} presents the experimental results, followed by some analysis. Finally, conclusion and discussions are given in Section~\ref{sec_conclusion}.
}

The main notations used are listed below.

\begin{table}[htpb]
  \centering
  {
  \begin{tabular}{lc}
    \toprule
    Definition
    & Notation
     \\
    \midrule
    Number of edge servers & $N$ \\ 
    Number of clients for each edge server & $J$ \\ 
    Global model of the cloud server & $\bm w $ \\ 
    Edge global model of $i$-th edge server & $\bm w_i $ \\ 
    Edge personalized model of $i$-th edge server & $\bm \varphi_i $ \\
    Local edge model of $i$-th edge server and $j$-th client & $\bm \varphi_{i,j} $ \\
    Local personalized model & $\bm \theta_{i,j} $ \\
    Expected loss over the data of the $i$-th server and $j$-th client & $\ell_{i,j}$ \\
    Training data of $j$-th client in $i$-th server & $Z_{i,j}$ \\
    Hyperparameter for sparsity & $\gamma_1$, $\gamma_2$ \\
    Hyperparameter for personalization & $\lambda_1$, $\lambda_2$ \\
    Hyperparameter for aggregation & $\beta$ \\
    Global training round & $T$ \\
    Local training round & $R$ \\
    Number of edges to aggregate & $S$ \\
    Learning rate for updating $\bm w_i$ & $\eta_1$ \\
    Learning rate for updating $\bm \theta_{i,j}$ & $\eta_2$ \\
    Expectation function & $\mathbb{E}(\cdot)$\\
    Same-order infinitesimal function & $\mathcal{O}(\cdot)$\\
    \bottomrule
  \end{tabular}}
\end{table}

{\color{black}
\section{Related Work}\label{sec_related_work}
Traditional machine learning methods require local training data to be uploaded to a central server for centralized training. However, in practical scenarios like training on mobile phone data, sensor data in the Internet of Things, and data from external companies like banks, uploading local data poses a significant privacy risk. Thus, participants are often hesitant to expose their local data during training. To address these issues, McMahan et al. proposed a federated learning framework and its model aggregation algorithm, FedAvg~\cite{mcmahan2017communication}. They also developed a network protocol for federated learning in 2019~\cite{bonawitz2019towards}. The framework involves multiple participants and a central server, where participants use their local data for training and upload model updates to a parameter server. The global model is obtained by aggregating these updates, enabling multi-party collaborative machine learning while protecting local data. Federated learning has been successfully applied in various scenarios, including smart apps such as Google Keyboard, Siri Speech Classifier, and QuickType Keyboard, as well as cross-silo applications such as drug discovery, financial risk prediction, and smart manufacturing.

In federated learning, data is supposed to be independent and identically distributed, meaning each user's mini-batch sample must be statistically the same as a sample drawn uniformly from the entire training data set of all users. However, due to differences in devices, participants, enterprises, and scenarios, statistical diversity often exists among users, resulting in non-i.i.d. data. Non-i.i.d. data includes skewed feature distribution, skewed label distribution, different features of the same label, and different labels of the same feature~\cite{hsieh2020non}. Recent research on non-i.i.d. data in federated learning has primarily focused on skewed label distribution, where different users have different label distributions. To construct non-i.i.d. training samples, researchers partition existing "flat" datasets based on labels~\cite{kairouz2021advances}. However, global models trained on non-i.i.d. datasets often struggle to generalize well~\cite{mansour2020three}. To address this problem, various personalized federated learning methods have been proposed, including local customization based methods~\cite{li2018federated,t2020personalized,huang2021personalized,fallah2020personalized,achituve2021personalized}, personalized layer based methods~\cite{arivazhagan2019federated}, knowledge distillation based methods~\cite{hinton2015distilling,lin2020ensemble}, multi-task learning based methods~\cite{smith2017federated}, and lifelong learning based methods~\cite{shoham2019overcoming}.

In particular, local customization based methods fine-tune the local model to customize a personalized model. 
Fedprox~\cite{li2018federated} achieves good personalization through $\ell_2$-norm regularization.
pFedMe~\cite{t2020personalized} can decouple personalized model optimization from the global-model learning in a bi-level problem stylized based on Moreau envelopes.
Per-FedAvg~\cite{fallah2020personalized} sets up an initial meta-model that can be updated effectively after one more gradient descent step.
FedAMP and HeurFedAMP~\cite{huang2021personalized} utilize the attentive message passing to facilitate more collaboration with similar customers.
pFedGP~\cite{achituve2021personalized} is proposed as a solution to PFL based on Gaussian processes with deep kernel learning.
The proposed $\texttt{sFedHP}$ is a local customization based personalized FL method, so we compare the performance of $\texttt{sFedHP}$ with other local customization based personalized FL methods, including ~\cite{li2018federated,t2020personalized,huang2021personalized,fallah2020personalized,achituve2021personalized}.

However, personalized federated learning algorithms often prioritize the performance of individual client models over the global model's performance, which contradicts the fundamental objective of federated learning to develop a high-quality global model and hinder the inclusion of new clients. Additionally, due to the need for frequent communication between clients and the cloud server to ensure convergence performance, FL suffers from high latency and limited bandwidth. As a result, we propose $\texttt{sFedHP}$, a communication-friendly personalized federated learning scheme that produces a high-performance global model.

}

\section{Sparse Federated Learning with Hierarchical Personalized Models (sFedHP)} \label{sec_2}

\subsection{{\texttt{sFedHP}}: Problem Formulation}

\begin{figure}[!tp]
	\centering

	\includegraphics[width=3.2in]{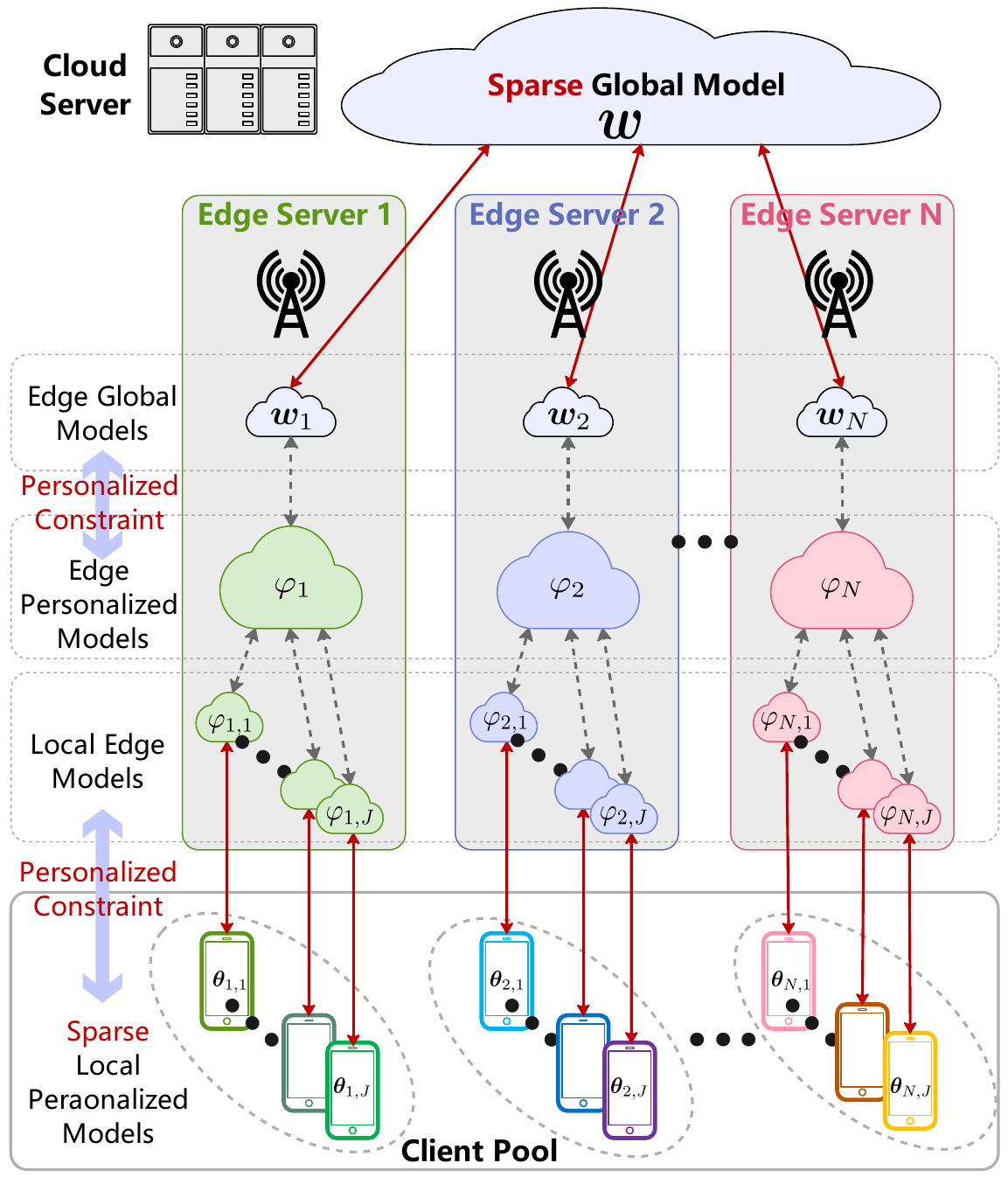}

	\caption{\color{black}Sparse client-edge-cloud FL with hierarchical personalized models.}
	\label{structure}
\end{figure}

Consider a client-edge-cloud framework with one cloud server, $N$ edge servers and $NJ$ clients, i.e., each edge server collects data from $J$ clients as showed in Fig.~\ref{structure}. In \texttt{sFedHP}, we aims to find a {\em sparse global-model} $\bm w$ based on local {\em sparse personalized models} $\bm \theta_{i,j},~i=1,...,N,~j=1,...,J$, which can greatly reduce the model size and communication costs, by minimizing
\begin{align}
\label{sFedPer-1}
    \text { {\texttt{sFedHP}}:}~ \min _{{\bm w} \in \mathbb{R}^{d}}\left\{F(\bm w)\triangleq \frac{1}{N} \sum_{i=1}^{N} F_{i}(\bm w)\right\},
\end{align}
with
\begin{align}
    F_i(\bm w)\triangleq & \frac{1}{J} \sum_{j=1}^{J} F_{i,j}(\bm w),\label{sFedPer-1.5}\\
    F_{i,j}(\bm w)=&\min _{\bm \varphi_{i,j}, \bm \theta_{i,j} \in \mathbb{R}^{d}}~\ell_{i,j}\left(\bm \theta_{i,j}\right)+\frac{\lambda_1}{2}\left\|\bm \theta_{i,j}-{\bm \varphi_{i,j}}\right\|_2^{2}\notag\\
	&\!+\!\gamma_1\phi_{\rho}(\bm \theta_{i,j}) \!+\! \frac{\lambda_2}{2}\left\|\bm \varphi_{i,j}-{\bm w}\right\|_2^{2}+\gamma_2\phi_{\rho}({\bm w}),\label{sFedPer-2}
\end{align}
where $\ell_{i,j}(\cdot): \mathbb{R}^{d} \rightarrow \mathbb{R}$ denotes the expected loss over the data distribution of the $i$-th edge server and the $j$-th client; $\lambda_1$ and $\lambda_2$ denoting regularization parameters that control the strength of ${\bm w}$ to the personalized model; $\bm \varphi_{i,j}$ is the local edge model of the $i$-th edge server and the $j$-th client; $\gamma_1$ and $\gamma_2$ denoting weight factors that control the sparsity level; and $\phi_{\rho}(\boldsymbol{\bm x})$ being a twice continuously differentiable approximation for $\|\bm x\|_1$~\cite{7755794}, which is given by
\begin{align}
\label{eq_varpi}
 \phi_{\rho}(\boldsymbol{\bm x})
    & =\rho \sum_{n=1}^{d} \log \cosh \left(\frac{{x}_{n}}{\rho}\right),
\end{align}
where $x_n$ denotes the $n$-th element in $\bm x$ and $\rho$ is a weight parameter, which controls the smoothing level. Note that we use $\phi_{\rho}({\cdot})$ instead of $\|\cdot\|_1$ to exploit the sparsity in ${\bm w}$ and ${\bm \theta}$.

{
\color{black}

{\texttt{sFedHP}}  utilizes the moreau envelope twice to partition the optimization problem into three stages, enabling independent optimization of the global model ${\bm w}$, client models ${\bm \theta_{i,j}}$, and edge server models ${\bm \varphi_{i,j}}$. The first application of the moreau envelope decouples the optimization of the global model from that of the edge server models, while the second separates the optimization of the client models from that of the edge server models. Leveraging the moreau envelope allows us to achieve personalized models for clients and edge servers while maintaining proximity to the reference model, resulting in enhanced performance on non-i.i.d data. To encourage sparsity in the global model ${\bm w}$, we permit clients to discover their own sparse models ${\bm \theta_{i,j}}$ using continuously differentiable approximated $\ell_1$-norm constraints within a reasonable distance from the reference point ${\bm w}$. 

In the first stage, the optimal local personalized model $\bm{\hat\theta}_{i,j}$ is obtained by minimizing \eqref{sFedPer-2} w.r.t. the data distribution of the $i$-th edge server and $j$-th client in the third stage as the following,
\begin{align}
\label{theta-1}
    \bm{\hat\theta}_{i,j}(\bm \varphi_{i,j}) = 
    \arg\min _{\bm \theta_{i,j} \in \mathbb{R}^{d}}
    L_{i,j}(\bm \theta_{i,j}) + 
    \frac{\lambda_1}{2}\left\|\bm \theta_{i,j}-{\bm \varphi_{i,j}}\right\|_2^{2},
\end{align}
where $L_{i,j}(\bm \theta_{i,j}) \triangleq \ell_{i,j}(\bm \theta_{i,j})+\gamma_1\phi_\rho(\bm \theta_{i,j})$.

In the second stage, the edge personalized model $\bm \varphi_{i}$ and the local edge model $\bm \varphi_{i,j}$ are determined by minimizing \eqref{sFedPer-2} w.r.t. the client models of the $i$-th edge server,
\begin{align}
\label{y-1}
\bm{\hat \varphi}_{i}&=\frac{1}{J} \sum_{i=1}^{J} \bm{\hat \varphi}_{ij},\\
\bm{\hat \varphi}_{ij}&= \arg\min _{\bm \varphi_{ij} \in \mathbb{R}^{d}}~\frac{\lambda_1}{2}\left\|\bm{\hat\theta}_{i,j}-{\bm \varphi_{ij}}\right\|_2^{2}+\frac{\lambda_2}{2}\left\|\bm \varphi_{ij}-{\bm w_i}\right\|_2^{2}, \notag
\end{align}

In the third stage, $\bm w$ is determined by utilizing the sparse model aggregation from multiple edges.   
}

{\color{black}
\begin{assumption}\label{assumption_1} (Strong convexity and smoothness)
Assume that $\ell(\bm w):\mathbb{R}^{d} \rightarrow \mathbb{R} $ is $(a)$ $\mu$-strongly convex or $(b)$ nonconvex and $L$-smooth on $\mathbb{R}^{d}$, then we respectively have the following inequalities 
\begin{align}
(a)~& \ell(\bm w) \geq \ell (\bm{\bar w})+ \langle\nabla \ell  (\bm{\bar w} ), \bm w - \bm{\bar w} \rangle+\frac{\mu}{2} \|\bm w - \bm{\bar w}  \|_2^{2},\notag \\
(b)~& \|\nabla \ell(\bm w)-\nabla \ell (\bm{\bar w})\|_2 \leq L\|\bm w-\bm{\bar w}\|_2. \notag
\end{align}
\end{assumption}

\begin{assumption}\label{assumption_2} (Bounded variance)
The variance of stochastic gradients (sampling noise) in each client is bounded by
\begin{align}
    \mathbb{E}_{\mathcal{Z}}\left[\left\|\nabla \tilde{\ell}_{i,j}\left(\bm w ; Z_{i,j}\right)-\nabla \ell_{i,j}(\bm w)\right\|^{2}_2\right] \leq \gamma_{\ell}^{2},\notag
\end{align}
where $Z_{i,j}$ is the training data randomly drawn from the distribution of client $i$ and edge server $j$.
\end{assumption}

\begin{assumption}\label{assumption_3} (Bounded diversity)
The diversity of client's data distribution is bounded by
\begin{align}
    \frac{1}{NJ} \sum_{i,j=1}^{N,J}\left\|\nabla \ell_{i,j}(\bm w)-\nabla \ell(\bm w)\right\|^{2}_2 \leq \sigma_{\ell}^{2}.\notag
\end{align}
\end{assumption}
While Assumption~\ref{assumption_1}, \ref{assumption_2} and \ref{assumption_3} are widely used in FL gradient calculation and convergence analysis~\cite{beck2017first,t2020personalized,karimireddy2020scaffold,fallah2020personalized}.
}

\subsection{\texttt{sFedHP:} Algorithm}
{\color{black}

The pseudocode for {\texttt{sFedHP}} is outlined in Algorithm~1. During the $t$-th communication round, the cloud server broadcasts the global model $\bm w^t$ to all edge servers. Subsequently, the edge servers and their clients conduct R rounds of iterative training and upload their edge global model $\bm w_i^t$ to aggregate a new global model $\bm w^{t+1}$. As previously mentioned, {\texttt{sFedHP}} decomposes the optimization process into three stages.

In the first level, the local personalized model $\bm{\hat \theta}_{i,j}$ of the $i$-th edge server and the $j$-th client is determined by solving  \eqref{theta-1}, whose parameters are sparse and can reduce the communication load between clients and edge servers. Note that \eqref{theta-1} can be easily solved by many first-order approaches, for example, Nesterov’s accelerated gradient descent, based on the gradient. However, calculate the exact $\nabla \ell_{i,j}(\bm \theta_{i,j})$ requires the distribution of $Z_{i,j}$, we hence use the unbiased estimate by sampling a mini-batch of data $\mathcal{D}_{i,j}$,
\begin{align}
   \nabla \check \ell_{i,j}(\bm\theta_{i,j},\mathcal{D}_{i,j}) = \frac{1}{|\mathcal{D}_{i,j}|}\sum_{Z_{i,j} \in \mathcal{D}_{i,j}} \nabla  \ell_{i,j}(\bm\theta_{i,j},Z_{i,j}).
\end{align}
such that $\mathbb{E}[ \nabla \check \ell_{i,j}(\bm\theta_{i,j},\mathcal{D}_{i,j}) ] = \nabla \ell_{i,j}(\bm \theta_{i,j})$. Thus, we solve the following minimization problem instead of solving \eqref{theta-1} to obtain an approximated local personalized model
\begin{align}
\label{H-min}
\bm{\check \theta}_{i,j}(\bm \varphi_{i,j}^{t,r}) = \arg\min _{\boldsymbol{\theta}_{i,j} \in \mathbb{R}^{d}} H(\bm \theta;\bm \varphi_{i,j}^{t,r},\mathcal{D}_{i,j}),
\end{align}
where $H(\bm \theta_{i,j};\bm \varphi_{i,j}^{t,r},\mathcal{D}_{i,j})=\check \ell_{i,j}\left(\boldsymbol{\theta}_{i,j},\mathcal{D}_{i,j} \right) +\gamma_1 \phi_{\rho}(\bm \theta_{i,j}) + \frac{\lambda_1}{2}\left\|\bm \theta_{i,j}-\bm \varphi_{i,j}^{t,r}\right\|^{2}_2$. Similarly, \eqref{H-min} can be solved by Nesterov’s accelerated gradient descent, we let the iteration goes until the condition
\begin{align}
\| \nabla H(\bm {\check\theta}_{i,j};\bm \varphi_{i,j}^{t,r},\mathcal{D}_{i,j}) \|^2_2 \leq \nu, 
\end{align}
is reached, where $\nu$ is an accuracy level.

In the second level, after clients' local personalized models are updated, the local edge models are determined by
\begin{align}
\label{y-3}
\bm{\check \varphi}_{i,j}^{t,r+1}&= (\lambda_1\bm{\check\theta}_{i,j}^{t,r} + {\lambda_2\bm w_{i}^{t,r}}) / (\lambda_1 + \lambda_2), 
\end{align}
where $\bm{\check \varphi}_{i}=\frac{1}{J} \sum_{i=1}^{J} \bm{\check \varphi}_{i,j}$ is the current edge peraonalized model of the $i$-th edge server.

In the third level, once edge personalized models are updated, the edge global models can be updated by stochastic gradient descent as follows
\begin{align}
\label{eq_delta_w_fij}
    \bm w_{i}^{t, r+1} = &~\bm w_{i}^{t, r}-\eta_{1} \nabla F_{i}\left(\bm w_{i}^{t, r}\right) , \\
    =&~ \bm w_{i}^{t,r} - \eta_1 [ \lambda_2(\bm w_{i}^{t,r} - \bm{\check \varphi}_{i}^{t,r+1}) +  \gamma_2\nabla\phi_{\rho}({\bm w_{i}^{t,r}}) ], \nonumber
\end{align}
where $\eta_1$ is a learning rate and ${\bm w_{i}^{t,r}}$ is the edge global model of $i$-th edge server at the global round $t$ and edge round~$r$. The global model ${\bm w^{t+1}}$ is determined by utilizing the edge global model ${\bm w_{i}^{t+1}}={\bm w_{i}^{t, R}}$ aggregation from multiple edges. Moreover, similarly with~\cite{t2020personalized,karimireddy2020scaffold}, an additional parameter $\beta$ is used for global-model update to improve the convergence performance.

Note that communication between the clients and edge servers is more efficient than between the clients and the cloud server. The latter's communication is high cost and latency because the distance is relatively long.

}

\begin{algorithm}[!tp] 
\label{Algorithm1}
{\color{black}
	\caption{{\texttt{sFedHP}}: Sparse Federated Learning with Hierarchical Personalization Models Algorithm}
	\begin{tabular}{l}
	\textbf{Input}: hyperparameters in \eqref{sFedPer-2} $\{\lambda_1, \lambda_2, \gamma_1, \gamma_2, \rho \}$ \\
	\hspace{0.95cm} communication ronds $\{T, R\}$ \\
	\hspace{0.95cm} init parameters $\bm w^{0}$\\
	\hspace{0.95cm} client learning rate $\eta_1$, server learning rate $\eta_2$ \\
{\bf{Cloud server executes:}} \\
	\hspace{0.3cm}\bf{for} $t=0,1,...,T-1$ \bf{do} \\
       	\hspace{0.6cm}\bf{for} $i=1,2,...,N$ \bf{in parallel do} \\
			\hspace{0.9cm}$\bm w_i^{t+1} \leftarrow \text{EdgeUpdate}(i,\bm w^t)$  \\
		\hspace{0.6cm}$\mathcal{S}^{t} \leftarrow \text{(random set of $S$ edge servers)}$ \\
		\hspace{0.6cm}$\bm w^{t+1}=(1-\beta) \bm w^{t}+\frac{\beta}{S} \sum_{i \in S^t} {\bm w_i^{t+1}}$ \\
	$\textbf{EdgeUpdate}(i,\bm w^t)\textbf{:}$ \\
		\hspace{0.3cm}$\bm \varphi_{i}^{t,0} = \bm w_i^{t,0} = \bm w^{t} $ \\
		\hspace{0.3cm}{\bf{for}} $r=0,1,...,R-1$ \bf{do}  \\
			\hspace{0.6cm}\bf{for} $j=1,...,J$ \bf{in parallel do}  \\
			\hspace{0.9cm}
			Update the local personalized model $\bm{\theta}_{i,j}$:\\
			\hspace{0.9cm}
		    $\mathcal{D}_{i,j} \leftarrow \text{(sample a mini-batch with size $|\mathcal{D}|$)}$ \\
		    \hspace{0.9cm}
		    $\bm{\check \theta}_{i,j}^{t,r}(\bm \varphi_{i}^{t,r}) = \arg\min _{\boldsymbol{\theta}_{i,j} \in \mathbb{R}^{d}} H(\bm \theta_{i,j};\bm \varphi_{i}^{t,r},\mathcal{D}_{i,j})$\\
            \hspace{0.9cm}
            Update the local edge model $\bm{ \varphi}_{i,j}$ according to \eqref{y-1}:\\
			\hspace{0.9cm}
			$\bm{\check \varphi}_{i,j}^{t,r+1}= {(\lambda_1\bm{\check\theta}_{i,j}^{t,r} + \lambda_2{\bm w_{i}^{t,r}}) } / {(\lambda_1 + \lambda_2)}$ \\
		\hspace{0.6cm} Update the edge personalized model $\bm{\varphi}_{i}$:\\
		\hspace{0.6cm} $\bm{\varphi}_{i}^{t,r+1}=\frac{1}{J}\sum_{j=1}^{J}\bm{\check \varphi}_{i,j}^{t,r+1}$\\		
		\hspace{0.6cm}$\bm w_{i}^{t,r+1}=\bm w_{i}^{t,r}-\eta_{1} [\lambda_2(\bm w_{i}^{t,r} \!-\! \bm{\varphi}_{i}^{t,r+1}) \!+\! \gamma_2\nabla\phi_{\rho}({\bm w_{i}^{t,r}})]$ \\
		\hspace{0.3cm}Return $\bm w_{i}^{t+1} = \bm w_{i}^{t,R}$ to the cloud server\\
	\end{tabular}
}
\end{algorithm}

\section{Convergence Analysis } \label{sec_3}

\subsection{Convergence Theorems}
{\color{black}
In this section, we present the convergence or \texttt{sFedHP}. We first prove Theorem~\ref{theorem_1} under Assumptions~\ref{assumption_1}-\ref{assumption_3}, which presents the smoothness and strong convexity proporties  of $F_{i,j}(\bm w)$. 
}

\begin{theorem}\label{theorem_1}
If $\ell_{i,j}$ is convex or nonconvex with L-Lipschitz $\nabla\ell_{i,j}$, then $F_{i,j}$ is $L_F$-smooth with $L_F=\lambda_2+\frac{\gamma_2}{\rho}$ (with the condition that $\lambda_2 > 4L+\frac{4\gamma_1}{\rho}$ for nonconvex $L$-smooth $\ell_{i,j}$) and
if $\ell_{i,j}$ is $\mu$-strongly convex, then $F_{i,j}$ is $\mu_F$-strongly convex with $\mu_F=\frac{\lambda_1\lambda_2\mu}{\lambda_1\mu+\lambda_2\mu+\lambda_1\lambda_2}$.
\end{theorem}

\begin{proof}
We first prove some interesting character of $\phi_{\rho}(\bm x)$ in \eqref{eq_varpi}, which is convex smooth approximation to $\|\bm x\|_1$~\cite{7755794} and
\begin{align}
    0\leq\nabla^2\phi_{\rho}(x)=\frac{1}{\rho}\left(1-\left(\tanh(x/\rho)\right)^2\right)\leq\frac{1}{\rho}\notag,
\end{align}
which yields $\nabla\phi_{\rho}(x)-\nabla\phi_{\rho}(\bar x) \leq \frac{1}{\rho}\left(x - \bar x\right)$, then we have
\begin{align}
 &~   \left\|\nabla\phi_{\rho}(\bm x)-\nabla\phi_{\rho}(\bm{\bar x})\right\|_2 \notag\\
=&~\sqrt{\sum_{n=1}^{d}\left(\tanh \left(x_{n} / \rho\right)-\tanh \left(\bar x_{n} / \rho\right)\right)^2}\notag\\
\leq &~\frac{1}{\rho}\sqrt{\sum_{n=1}^{d}\left(x_{n} - \bar x_{n}\right)^2} = \frac{1}{\rho}\left\|\bm x-\bm{\bar x}\right\|_2.\notag
\end{align}

Hence, $\phi_{\rho}$ is convex function with $\frac{1}{\rho}$-Lipschitz $\nabla\phi_\rho$. 

Let $L_{i,j}(\bm \theta_{i,j})=\ell_{i,j}(\bm \theta_{i,j})+\gamma_1\phi_\rho(\bm \theta_{i,j})$. On the one hand if $\ell_{i,j}$ is $\mu$-strong convex. For $\alpha \in [0,1]$, we hence have 
\begin{align}
\label{eq_strong_convex_ell}
	&~\ell_{i,j}\left(\alpha \bm \theta_{i,j} + (1-\alpha)\bm{\bar \theta}_{i,j}\right)  \\
\leq&~\alpha \ell_{i,j}(\bm \theta_{i,j})+(1-\alpha)\ell_{i,j}(\bm{\bar \theta}_{i,j})-\frac{\mu}{2}\alpha(1-\alpha)\|\bm \theta_{i,j}-\bm {\bar\theta}_{i,j}\|_2. \notag
\end{align}

In addition, by noting that $\phi_\rho$ is convex function, we  have
\begin{align}
\label{eq_convex_phi}
	\phi_\rho\left(\alpha \bm \theta_{i,j} + (1-\alpha)\bm {\bar \theta}_{i,j}\right) \leq \alpha \phi_\rho(\bm\theta_{i,j})+(1-\alpha)\phi_\rho(\bm {\bar \theta}_{i,j}).
\end{align}

Combining \eqref{eq_strong_convex_ell} and \eqref{eq_convex_phi} yields
\begin{align}
	&~L_{i,j}\left(\alpha \bm \theta_{i,j} + (1-\alpha)\bm {\bar \theta}_{i,j} \right) \notag \\
\leq &~ \alpha L_i(\bm \theta_{i,j})+(1-\alpha)L_i(\bm {\bar \theta}_{i,j})-\frac{\mu}{2}\alpha(1-\alpha)\|\bm \theta_{i,j}-\bm {\bar \theta}_{i,j}\|_2,  \notag
\end{align}
which shows that  if $\ell_{i,j}$ is $\mu$-strong convex, $L_{i,j}(\bm \theta_{i,j})=\ell_{i,j}(\bm \theta_{i,j})+\gamma_1\phi_\rho(\bm \theta_{i,j})$ is $\mu$-strong convex.

On the other hand, if $\ell_{i,j}$ is nonconvex with $L$-Lipschitz $\nabla\ell_{i,j}$, we have
\begin{align}
	&~\|\nabla L_{i,j}(\bm \theta_{i,j})-\nabla L_{i,j}(\bm {\bar \theta}_{i,j})\|_2\notag\\
	=&~\|\nabla \ell_{i,j}(\bm \theta_{i,j})-\nabla \ell_{i,j}(\bm {\bar \theta}_{i,j})+\gamma_1\nabla\phi_{\rho}(\bm \theta_{i,j})-\gamma_1\nabla\phi_{\rho}(\bm {\bar \theta}_{i,j})\|_2\notag\\
	\leq &~ \|\nabla \ell_{i,j}(\bm \theta_{i,j})-\nabla \ell_{i,j}(\bm {\bar \theta}_{i,j})\|_2+\gamma_1\|\nabla\phi_{\rho}(\bm \theta_{i,j})-\nabla\phi_{\rho}(\bm {\bar \theta}_{i,j})\|_2\notag\\
	\leq&~ (L+\frac{\gamma_1}{\rho}) \left\|\bm w-\bm{\bar w}\right\|_2\notag,
\end{align}
which shows that  if $\ell_{i,j}$ is $L$-smooth, then $L_{i,j}(\bm \theta_{i,j})=\ell_{i,j}(\bm \theta_{i,j})+\gamma_1\phi_\rho(\bm \theta_{i,j})$ is $L+\frac{\gamma_1}{\rho}$-smooth.

Let $g(\bm y_{i,j})=M_{L_{i,j}}^{\bm\theta_{i,j}}(\bm y_{i,j})=\min_{\bm\theta_{i,j}\in\mathbb{R}^d}~L_{i,j}(\bm\theta_{i,j})+\frac{\lambda_1}{2}\|\bm y_{i,j}-\bm\theta_{i,j}\|_2^2$, which is the famous Moreau envelope~\cite{beck2017first}, thus if $L_{i,j}$ is convex or nonconvex but $(L+\frac{\gamma_1}{\rho})$-smooth, then $g(\bm y_{i,j})$ is $L_g$-smooth with $L_g=\lambda_1$(with the condition that $\lambda_1>2(L+\frac{\gamma_1}{\rho})$ for nonconvex $L+\frac{\gamma_1}{\rho}$-smooth ). Therefore if $L_{i,j}$ is $\mu$-strong convex, $g(\bm y_{i,j})$ is $\mu_g$-strong convex wiht $\mu_g=\frac{\lambda_1\mu}{\lambda_1+\mu}$.
	
Let $G_{i,j}(\bm w)=M_{g_{i,j}}^{\bm y_{i,j}}(\bm y_{i,j})=\min_{\bm y_{i,j}\in\mathbb{R}^d}~g_{i,j}(\bm y_{i,j})+\frac{\lambda_2}{2}\|\bm w-\bm y_{i,j}\|_2^2$, similarly we have if $g_{i,j}$ is convex or nonconvex with $L_g$-smooth, then $G_{i,j}(\bm w)$ is $L_G$-smooth with $L_G=\lambda_2$ (with the condition that $\lambda_2>2L_g$ for nonconvex $L_g$-smooth). Therefore if $g_{i,j}$ is $\mu_g$-strong convex, yeilds $G_{i,j}(\bm w)$ is $\mu_G$-strong convex with 
\begin{align}
\mu_G=\frac{\lambda_2\mu_g}{\lambda_2+\mu_g}=\frac{\lambda_1\lambda_2\mu}{\lambda_1\mu+\lambda_2\mu+\lambda_1\lambda_2}. \notag
\end{align}

Furthermore, recall that
\begin{align}
    F_{i,j}(\bm w)&=\min _{\bm y_{i,j}, \bm \theta_{i,j} \in \mathbb{R}^{d}}~\ell_{i,j}\left(\bm \theta_{i,j}\right)+\frac{\lambda_1}{2}\left\|\bm \theta_{i,j}-{\bm y_{i,j}}\right\|_2^{2}\notag\\
	&~~~~+\gamma_1\phi_{\rho}(\bm \theta_{i,j})+\frac{\lambda_2}{2}\left\|\bm y_{i,j}-{\bm w}\right\|_2^{2}+\gamma_2\phi_{\rho}({\bm w})\notag\\
	&=\min_{\bm y_{i,j}\in\mathbb{R}^d}~g_{i,j}(\bm y_{i,j})+\frac{\lambda_2}{2}\|\bm w-\bm y_{i,j}\|_2^2+\gamma_2\phi_\rho(\bm w)\notag\\
	&=G(\bm w)+\gamma_2\phi_\rho(\bm w) .\notag
\end{align}

Using the character of $\phi_\rho$ mentioned before, we have the similarly couclusion that if $G_i$ is $\mu_G$-strongly convex, $F_{i,j}$ is  $\mu_F$-strong convex with $\mu_F=\mu_G=\frac{\lambda_1\lambda_2\mu}{\lambda_1\mu+\lambda_2\mu+\lambda_1\lambda_2}$, and if $G_i$ is $L_G$-smooth, $F_{i,j}$ is $L_F$-smooth with $L_F=L_G+\frac{\gamma_2}{\rho}=\lambda_2+\frac{\gamma_2}{\rho}$. Finally, combining the first-order condition $\lambda_1(\hat{\bm y}_{i,j}-\hat{\bm \theta}_{i,j})+\lambda_2(\hat{\bm y}_{i,j}-\bm w)=0$, we complete the proof.
\end{proof}

For unique solution $\bm{w}^{*}$ to \texttt{sFedHP}, we have the following convergence theorems, Theorems~\ref{theorem_2} and \ref{theorem_3}, based on Assumptions~\ref{assumption_1}-\ref{assumption_3} and Theorem~\ref{theorem_1}. 

\begin{theorem}\label{theorem_2}
(Strongly convex \texttt{sFedHP}'s convergence). Let Assumptions~\ref{assumption_1}(a) and~\ref{assumption_2} hold, there exists an $\tilde\eta \leq \min[\frac{1}{\mu_F}, \frac{1}{3L_F(3+128L_F/(\mu_F\beta))}]$, such that
\begin{align}
	(a)&\frac{1}{T}\sum_{t=0}^{T-1} \mathbb{E}\left[ F_{}\left(\bm w^{t}\right)- F_{}\left(\bm w^{*}\right)\right]\leq \frac{\Delta_0}{\tilde\eta T}+M_1 +\tilde\eta M_2, \notag\\
    (b)&\frac{1}{NJ} \sum_{i,j=1}^{NJ} \mathbb{E}\left[\left\|\tilde{\theta}_{i,j}^{T}\left(\bm w^{T}\right)-\bm w^{*}\right\|_2^{2}\right] \notag\\
    &\leq\mathcal{O}\left(\frac{L_F+\lambda^2}{\lambda^2\mu_{F}}\right) \mathbb{E}\left[F\left({\bm w}_{T}\right)-F^{*}\right]+\mathcal{O}\left(D_1\right) , \notag
\end{align}
where $\Delta_{0}\triangleq \left\|\bm w_{0}-\bm w^{*}\right\|^{2}$, $M_1=\frac{8\delta^2\bar\lambda^2}{\mu_F}$, $M_2=6(1+64\frac{L_F}{\mu_F\beta})\sigma_{F,1}^2 + \frac{128L_F\delta^2\bar\lambda^2}{\mu_F\beta}$, 
$\lambda \triangleq \frac{\lambda_1\lambda_2}{\sqrt{\lambda_1^2+\lambda_2^2}}$, $\bar\lambda \triangleq \frac{\lambda_1\lambda_2}{\lambda_1+\lambda_2}$ and $D_1={\sigma_{F, 1}^{2}}/{\lambda^{2}}+{\gamma_2^{2}d_s^2}/{\bar\lambda^2}+\delta^{2}$ with $\sigma_{F, 1}^{2} \triangleq   \frac{1}{NJ} \sum_{i,j=1}^{N,J}\left\|\nabla F_{i,j}\left(\bm w^{*}\right)\right\|_2^{2}$, $\delta$ being the sampling noise and $d_s$ denoting the number of non-zero elements in the global-model $\bm w$.
\end{theorem}

\begin{theorem}\label{theorem_3}
(Nonconvex and smooth \texttt{sFedHP}’s convergence). Let Assumptions~\ref{assumption_1}(b), ~\ref{assumption_2} and \ref{assumption_3} hold, there exists an $\tilde\eta \leq \min[\frac{1}{588L_F\lambda^{2})},\frac{\beta}{2L_F}]$ with $\lambda\leq\sqrt{16(L+\frac{\gamma_1}{\rho})^2+1}$, such that
\begin{align}
	(a)&\frac{1}{T}\sum_{t=0}^{T-1}\mathbb{E}\left[\left\|\nabla F\left(\bm w^{t}\right)\right\|_2^{2}\right]\leq4(\frac{\Delta_F}{\tilde\eta T}+ G_1 + \tilde{\eta}^{1}G_2)\notag\\
    (b)&\frac{1}{T}\sum_{t=0}^{T-1}\frac{1}{NJ} \sum_{i,j=1}^{N} \mathbb{E}\left[\left\|\tilde{\theta}_{i,j}^{}\left(\bm w^{t}\right)-\bm w^{t}\right\|_2^{2}\right]\notag\\
    &~~~~~\leq\frac{1}{T}\sum_{t=0}^{T-1}\mathbb{E}\left[\left\|\nabla F\left(\bm w^{t}\right)\right\|_2^{2}\right]+\mathcal{O}\left(D_2\right), \notag
\end{align}
where $\Delta_{F}\triangleq F\left(\bm w_{0}\right)-F^{*}$, $G_1 = 3\beta\bar\lambda^{2} \delta^{2}$, $G_2 = \frac{3L_F\left(49 R \sigma_{F, 2}^{2}+16 \delta^{2}\bar\lambda^{2}\right)}{R}$ and $D_2={\sigma_{F, 2}^{2}}/{\lambda^{2}}+{\gamma_2^{2}d_s^2}/{\bar\lambda^2}+\delta^{2}$ with $\sigma_{F, 2}^{2} \triangleq  \frac{\lambda^2}{\lambda^2-16(L+\frac{\gamma_1}{\rho})^2}\sigma_{\ell}^2$.
\end{theorem}

Theorems~\ref{theorem_2}(a) and~\ref{theorem_3}(a) show the convergence results of the global-model. $\Delta_{0}$ and $\Delta_{F}$ denote the initial error which can be reduced linearly. Theorem~\ref{theorem_2}(b) and~\ref{theorem_3}(b)  show that the convergence of personalized client models in average to a ball of center $\bm w^*$ and radius $\mathcal{O}(D_1)$ and $\mathcal{O}(D_2)$, respectively, for $\mu$-strongly convex and nonconvex. Note that, our \texttt{sFedHP} obtain sparsity with a convergence speed cost $\mathcal{O}({\gamma_2^{2}d_s^2}/{\bar\lambda^2})$, the experimental results show that a good sparsity can be obtained by only using a tiny $\gamma_2$, yields ${\gamma_2^{2}d_s^2}/{\bar\lambda^2} \ll \{ {\sigma_{F, 1}^{2}}/{\lambda^{2}}, {\sigma_{F, 2}^{2}}/{\lambda^{2}}\}$, i.e., the convergence speed cost of the sparse constraints can be omitted. 
{\color{black}Note that, according to our experiments, limiting the learning rate does not increase the training time. This technique is both useful and common in Federated Learning analysis. Similar to the works in ~\cite{t2020personalized, karimireddy2020scaffold}, we have applied this technique to our theorem analysis.}
The proofs of Theorems~\ref{theorem_2} and~\ref{theorem_3} are presented in Appendix.

\subsection{Some Important Lemmas}
In this subsection, we present some important lemmas, which are well used in the proof of Theorems~\ref{theorem_2} and \ref{theorem_3} ,to help better understand the conclusion of the these convergence theorem.

We re-write the local update in \eqref{eq_delta_w_fij} as follow:
\begin{align}
    \bm w_{i}^{t}=\bm w_{i}^{t,r}-\eta_1 \underbrace{\lambda_2\left[\left(\bm w_{i}^{t,r}-\tilde{\bm y}_{i}^{t,r+1}\right)+\gamma_2\nabla\phi_\rho(\bm w_{i}^{t,r})\right]}_{\triangleq  g_{i}^{t,r}},\notag
\end{align}
which implies
\begin{align}
    \eta_1 \sum_{r=0}^{R-1} g_{i}^{t,r}=&~\sum_{r=0}^{R-1}\left(\bm w_{i}^{t,r}-\bm w_{i, r+1}^{t}\right) 
    =\bm w^{t}-\bm w_{i}^{t,R},\notag
\end{align}
where $g_{i}^{t,r}$ can be interpreted as the biased estimate of $\nabla F_{i}(\bm w^{t,r}_i)$ since $\mathbb{E}(g_{i}^{t,r}) \neq  \nabla F_{i}(\bm w^{t,r}_i)$.
Then, we re-write the global update as
\begin{align}
    \bm w^{t+1} 
    &=\bm w^{t}-\underbrace{\eta_1 \beta R}_{\triangleq  \tilde{\eta}} \underbrace{\frac{1}{S R} \sum_{i \in \mathcal{S}^{t}} \sum_{r=0}^{R-1} g_{i}^{t,r}}_{\triangleq  g^{t}},\notag
\end{align}
where $\tilde{\eta}$ and $g^{t}$ can be respectively considered as the step size and approximate stochastic gradient of the global update, which cause drift error in one-step update of the global model formulated in Lemma~\ref{lemma_3}.

\begin{lemma}[One-step global update]
\label{lemma_3}
Let Assumption 1(b) holds. We have
\begin{align}
    &\mathbb{E}\left[\left\|\bm w^{t+1}-\bm w^{*}\right\|_2^{2}\right] \notag\\
    &\leq \mathbb{E}\left[\left\|\bm w^{t}-\bm w^{*}\right\|_2^{2}\right]-\tilde{\eta}\left(2-6 L_{F} \tilde{\eta}\right)\mathbb{E}\left[F\left(\bm w^{t}\right)-F\left(\bm w^{*}\right)\right]\notag\\
    &~~~~~~+\frac{\tilde{\eta}\left(3 \tilde{\eta}+1 / \mu_{F}\right)}{N R} \sum_{i, r}^{N, R} \mathbb{E}\left[\left\|g_{i}^{t,r}-\nabla F_{i}\left(\bm w^{t}\right)\right\|^{2}_2\right]\notag\\
    &~~~~~~+3 \tilde{\eta}^{2}\frac{1}{NJ} \sum_{i,j=1}^{NJ}\mathbb{E}\left[\left\|\nabla F_{i,j}\left(\bm w^{t}\right)-\nabla F\left(\bm w^{t}\right)\right\|_2^{2}\right],\notag
\end{align}
where $\sum_{i, r}^{N, R}$ and $\sum_{i,j}^{NJ}$ are respectively used as alternatives for $\sum_{i=1, r=0}^{N, R-1}$ and $\sum_{i,j = 1}^{NJ}$.
\end{lemma}

The various parts of Lemma~\ref{lemma_3} are discussed in detail in the following Lemmas.

\begin{lemma}[Bounded diversity of $\bm \theta_{i,j}$ w.r.t. mini-batch sampling]
\label{lemma_1}
For the solution $\bm{\tilde \theta}_{i,j}(\bm y_{i,j}^{t,r})$ to \eqref{H-min},  we have
\begin{align}
        &\mathbb{E}\left[\left\|\bm{\tilde{\theta}}_{i,j}\left(\bm y_{i,j}^{t,r}\right) - \bm{\hat{\theta}}_{i,j}\left(\bm y_{i,j}^{t,r}\right)\right\|_2^{2}\right] \leq \delta^{2},\notag
\end{align}
with
\begin{align}
\delta^{2} \triangleq  \left\{\begin{array}{ll}
\frac{2}{(\lambda_1+\mu)^{2}}\left(\frac{\gamma_{l}^{2}}{|\mathcal{D}|}+\nu\right), & \text {if $A.\ref{assumption_1}(a)$ holds;} \\
\frac{2}{(\lambda_1-L-\frac{\gamma_1}{\rho})^{2}}\left(\frac{\gamma_{l}^{2}}{|\mathcal{D}|}+\nu\right), & \text {if $A.\ref{assumption_1}(b)$ holds. }
\end{array}\right.\notag
\end{align}
where $\lambda_1>L+\gamma_1/\rho$ is required for$A.\ref{assumption_1}(b)$ case.
\end{lemma}

\begin{lemma}[Bounded edge server drift error]
\label{lemma_4}
If $\tilde \eta\leq\frac{\beta}{2L_F}$, we have
\begin{align}
    &~~~~\frac{1}{N R} \sum_{i, r=1}^{N, R} \mathbb{E}\left[\left\|g_{i}^{t,r}-\nabla F_{i}\left(\bm w^{t}\right)\right\|_2^{2}\right] \leq 2 \bar\lambda^{2} \delta^{2}+ \notag\\
    &\frac{16 L_{F} \tilde{\eta}}{\beta}\left(3 \frac{1}{NJ} \sum_{i,j=1}^{NJ} \mathbb{E}\left[\left\|\nabla F_{i,j}\left(\bm w^{t}\right)\right\|_2^{2}\right]+\frac{2 \bar\lambda^{2} \delta^{2}}{R}\right), \notag
\end{align}
where  $g_{i}^{t,r} = \lambda_2\left[\left(\bm w_{i}^{t,r}-\tilde{\bm y}_{i}^{t,r+1}\right)+\gamma_2\nabla\phi_\rho(\bm w_{i}^{t,r})\right]$ and $\bar \lambda=\frac{\lambda_1\lambda_2}{\lambda_1+\lambda_2}$.
\end{lemma}

\begin{lemma}[Bounded diversity of $F_i$ w.r.t. distributed training]
\label{lemma_2}
Let $\bm w^{*}$ denote the optimal solution for \eqref{sFedPer-1}, we have
\begin{align}
        &\frac{1}{NJ} \sum_{i=1,j=1}^{N,J}\left\|\nabla F_{i,j}(\bm w)-\nabla F(\bm w)\right\|_2^{2} \leq  \notag \\
&\left\{\begin{array}{ll}4 L_{F}\left(F(\bm w)-F\left(\bm w^{*}\right)\right)+2\sigma_{F, 1}^{2},& \text {if $A.\ref{assumption_1}(a)$ holds;} \\
        \frac{16(L+\frac{\gamma_1}{\rho})^2}{\lambda^2-16(L+\frac{\gamma_1}{\rho})^2}\|\nabla F(\bm w)\|_2^{2}+2\sigma_{F, 2}^{2},& \text {if $A.\ref{assumption_1}(b)$ holds.}
\end{array}\right. \notag
\end{align}
where $\lambda > 4(L+{\gamma_1}/{\rho})$ is required for$A.\ref{assumption_1}(b)$ case,  $\sigma_{F, 1}^{2} \triangleq   \frac{1}{NJ} \sum_{i,j=1}^{N,J}\left\|\nabla F_{i,j}\left(\bm w^{*}\right)\right\|_2^{2}$, $\lambda = \frac{\lambda_1\lambda_2}{\sqrt{\lambda_1^2+\lambda_2^2}}$ and $\sigma_{F, 2}^{2} \triangleq  \frac{\lambda^2}{\lambda^2-16(L+\frac{\gamma_1}{\rho})^2}\sigma_{\ell}^2$.
\end{lemma}

Lemma~\ref{lemma_1} and Lemma~\ref{lemma_4} show the diversity and drift errors caused by mini-batch training strategy are bounded. While,  Lemma~\ref{lemma_2} shows the effection w.r.t. distributed training is bounded. By combining Lemma~\ref{lemma_3}-\ref{lemma_2}, we can get the recurrence relation between $\bm{w}^{t}-\bm{w}^{*}$ and $F(\bm{w}^{t})-F(\bm{w}^{*})$ in Theorems~\ref{theorem_2} and ~\ref{theorem_3}. The proofs of Lemma~\ref{lemma_3}-\ref{lemma_2} are presented in Appendix.

\section{Experimental Results} \label{sec_results}

\subsection{Performance Comparison}

To empirically highlight the performance of the proposed method, we first compare  {\texttt{sFedHP}} in non-saprsity setting with FedAvg~\cite{mcmahan2017communication}, hierarchical FedAvg (HierFAVG)~\cite{liu2020client} and local customization personalized FL methods, including
Fedprox~\cite{li2018federated},
Per-FedAvg~\cite{fallah2020personalized},
pFedMe~\cite{t2020personalized},
HeurFedAMP~\cite{huang2021personalized},
and pFedGP~\cite{achituve2021personalized} based on MNIST~\cite{lecun1998gradient}, Fashion-MNIST (FMNIST)~\cite{xiao2017fashion} and CIFAR-10~\cite{krizhevsky2009learning} datasets. For non-i.i.d. setups, we follow the strategy in~\cite{t2020personalized} for $N=20$ clients and assign each client a unique local data with only 5 out of 10 labels. For client-edge-cloud framework, we set 4 edge servers and each edge server manages 5 clients. All 20 clients are selected to generate the global model in following experiments.

Furthermore, we use two different dataset split settings to validate the performance of the above algorithms. In Setting 1, we used 200, 200, 100 training samples and 800, 800, 400 test samples in each class for MNIST, FMNIST and CIFAR-10 datasets respectively. In Setting 2, 900, 900, 450 training samples and 300, 300, 150 test samples are applied on MNIST, FMNIST and CIFAR-10 datasets for each class. We use the DNN model with two hidden layers of size [500, 200] for MNIST/FMNIST datasets and the VGG model for CIFAR-10 dataset with ``[16, `M', 32,  `M', 64 `M', 128,  `M', 128,  `M']" cfg setting. Each experiment is run at least 3 times to obtain statistical reports.

\begin{table}[bp] 
\centering
{
\color{black}
\caption{\color{black}The global model performance of different algorithms on MNIST, FMNIST, and CIFAR-10. We maintain $|\mathcal{D}|=20$, $T=800$, $K=5$, and $\eta_1=\eta_2=0.05$ across all algorithms and fine-tune other hyperparameters. Best results are bolded.}
  \label{dif-alg-table}
  \scalebox{1.0}{
  \begin{tabular}{cccc}
    \toprule
    \multirow{1}{*}{Dataset}  & \multirow{1}{*}{Method}  & \multicolumn{1}{c}{Setting 1  (\%)}  & \multicolumn{1}{c}{Setting 2 (\%)}\\
    \midrule
    \multirow{5}{*}{MNIST} 
       & FedAvg~\cite{mcmahan2017communication}     & 91.63$\pm0.12$ & 94.31$\pm0.08$\\
       & HierFAVG~\cite{liu2020client}     & 93.22$\pm0.03$ &  95.05$\pm0.07$\\
       & Fedprox~\cite{li2018federated}       & 93.43$\pm0.15$  & 90.04$\pm0.08$\\
       & pFedMe~\cite{t2020personalized}       & 89.34$\pm0.04$  & 93.04$\pm0.14$\\
       & {\color{black}sFedHP (Ours)}       & \textbf{94.93$\pm0.03$}  & \textbf{97.66$\pm0.06$} \\
    \midrule
    \multirow{5}{*}{FMNIST} 
       & FedAvg~\cite{mcmahan2017communication}     & 79.11$\pm0.09$  & 84.48$\pm0.06$\\
       & HierFAVG~\cite{liu2020client}     & 83.55$\pm0.15$  & 85.48$\pm0.08$\\
       & Fedprox~\cite{li2018federated}       & 78.67$\pm0.04$  & 84.33$\pm0.10$\\
       & pFedMe~\cite{t2020personalized}       & 80.57$\pm0.07$  & 85.14$\pm0.11$\\
       & {\color{black}sFedHP (Ours)}       & \textbf{85.64$\pm0.04$}  & \textbf{89.29$\pm0.05$} \\
    \midrule
    \multirow{5}{*}{CIFAR-10}
        & FedAvg~\cite{mcmahan2017communication}     & 42.36$\pm0.19$ & 81.74$\pm0.17$\\
        & HierFAVG~\cite{liu2020client}     & 52.76$\pm0.21$  & 87.28$\pm0.13$\\
       & Fedprox~\cite{li2018federated}     & 39.10$\pm0.23$  & 72.98$\pm0.35$\\
       & pFedMe~\cite{t2020personalized}     & 58.66$\pm0.15$  & 90.31$\pm0.22$\\
       & {\color{black}sFedHP (Ours)}       & \textbf{78.44$\pm0.08$}  & \textbf{94.09$\pm0.10$} \\
    \bottomrule
  \end{tabular}}
}  
\end{table}

\begin{table}[htbp]
\caption{Results of all of algorithms on On MNIST with setting 1.}
  \label{detail-result-dnn3-1}
  \centering
  \scalebox{0.85}{
  \begin{tabular}{ccccccccc}
    \toprule
    \multirow{2}{*}{Method} & \multirow{2}{*}{$\eta_1$} & \multirow{2}{*}{$\eta_2$} & \multirow{2}{*}{$\lambda$} & \multirow{2}{*}{$ K $}  & \multirow{2}{*}{$ E $} & \multirow{2}{*}{$\alpha $} &  \multicolumn{2}{c}{Acc.(\%)}\\
    \cmidrule(r){8-9}
        &   & & & & & &{PM}  &{GM}\\
    \midrule
        \multirow{1}{*}{Fedavg~\cite{mcmahan2017communication}}
        & 0.05  & - & - & 5 & - & - & - & 91.63 \\
    \midrule
        \multirow{1}{*}{HierFAVG~\cite{liu2020client}}
        & 0.05  & - & - & 5 & 4 & - & - & 93.22 \\
    \midrule
        \multirow{2}{*}{HeurFedAMP~\cite{huang2021personalized}}
        & 0.05  & 0.05 & - & 5 & - & 0.1 & 97.11 & -  \\
        & \textbf{0.05}  & \textbf{0.05}  & \textbf{-} & \textbf{5} & \textbf{-} & \textbf{0.5} & \textbf{98.55} & - \\
    \midrule
        \multirow{1}{*}{Fedprox~\cite{li2018federated}}
        & 0.05  & - & 0.001 & 5 & - & - & - & 93.43 \\
    \midrule
        \multirow{1}{*}{PerFedavg~\cite{fallah2020personalized}}
        & 0.05  & - & - & 5 & - & - & 56.5 & - \\
    \midrule
        \multirow{5}{*}{pFedMe~\cite{t2020personalized}}
        & 0.05  & 0.05 & 5 & 5 & - & - & {98.8} & {88.83} \\
        & \textbf{0.05}  & \textbf{0.05} & \textbf{15} & \textbf{5} & \textbf{-} & \textbf{-} & \textbf{98.92} & \textbf{89.34} \\
        &{0.05}  &{0.05}  &{20} &{5} &{-} &{-} & {98.91} & {89.06} \\
        & 0.05  & 0.05 & 25 & 5 & - & - & {97.6} & {80.12} \\
        & 0.05  & 0.05 & 30 & 5 & - & - & 52.07 & {10.41} \\
    \midrule
        \multirow{1}{*}{pFedGP~\cite{achituve2021personalized}}
        & 0.05 & - & - & 5 & - & - & 98.12 & - \\
    \midrule
        \multirow{5}{*}{\color{black}sFedHP (Ours)}
        & 0.05  & 0.05 & 5 & 5 & 4 & - & 98.24 & 93.06 \\
        & 0.05  & 0.05 & 15 & 5 & 4 & - & 96.55 & 91.85 \\
        & \textbf{0.05}  & \textbf{0.05}  & \textbf{20} & \textbf{5} & \textbf{4} & \textbf{-} & \textbf{97.83} & \textbf{94.93} \\
        & 0.05  & 0.05 & 25 & 5 & 4 & - & 96.63 & 95.39 \\
        & 0.05  & 0.05 & 30 & 5 & 4 & - & 96.23 & 95.63 \\
    \bottomrule
  \end{tabular}}
\end{table}

\begin{table}[htbp]
\caption{Results of all of algorithms on MNIST with setting 2.}
  \label{detail-result-dnn3-2}
  \centering
  \scalebox{0.85}{
  \begin{tabular}{ccccccccc}
    \toprule
    \multirow{2}{*}{Method} & \multirow{2}{*}{$\eta_1$} & \multirow{2}{*}{$\eta_2$} & \multirow{2}{*}{$\lambda$} & \multirow{2}{*}{$ K $}  & \multirow{2}{*}{$ E $} & \multirow{2}{*}{$\alpha $} &  \multicolumn{2}{c}{Acc.(\%)}\\
    \cmidrule(r){8-9}
        &   & & & & & &{PM}  &{GM}\\
    \midrule
        \multirow{1}{*}{Fedavg~\cite{mcmahan2017communication}}
        & 0.05  & - & - & 5 & - & - & - & 94.31 \\
    \midrule
        \multirow{1}{*}{HierFAVG~\cite{liu2020client}}
        & 0.05  & - & - & 5 & 4 & - & - & 95.05 \\
    \midrule
        \multirow{2}{*}{HeurFedAMP~\cite{huang2021personalized}}
        & 0.05  & 0.05 & - & 5 & - & 0.1 & 97.65 & - \\
        & \textbf{0.05}  & \textbf{0.05}  & \textbf{-} & \textbf{5} & \textbf{-} & \textbf{0.5} & \textbf{98.8} & - \\
    \midrule
        \multirow{1}{*}{Fedprox~\cite{li2018federated}}
        & 0.05  & - & 0.001 & 5 & - & - & - & 90.04 \\
    \midrule
        \multirow{1}{*}{PerFedavg~\cite{fallah2020personalized}}
        & 0.05  & - & - & 5 & - & - & 78.15 & - \\
    \midrule
        \multirow{5}{*}{pFedMe~\cite{t2020personalized}}
        & 0.05  & 0.05 & 5 & 5 & - & - &{99.17} &{91.14} \\
        & 0.05  & 0.05 & 15 & 5 & - & - & 99.38 &{91.60} \\
        & \textbf{0.05}  & \textbf{0.05}  & \textbf{20} & \textbf{5} & \textbf{-} & \textbf{-} & \textbf{99.39} & \textbf{93.04} \\
        & 0.05  & 0.05 & 25 & 5 & - & - & {98.72} & {89.47} \\
        & 0.05  & 0.05 & 30 & 5 & - & - & 52.08 & {10.53} \\
    \midrule
        \multirow{1}{*}{pFedGP~\cite{achituve2021personalized}}
        & 0.05 & - & - & 5 & - & - & 99.18 & - \\
    \midrule
        \multirow{5}{*}{\color{black}sFedHP (Ours)}
        & 0.05  & 0.05 & 5 & 5 & 4 & - & 99.01 & 94.69 \\
        & 0.05  & 0.05 & 15 & 5 & 4 & - & 98.04 & 94.06 \\
        & \textbf{0.05}  & \textbf{0.05}  & \textbf{20} & \textbf{5} & \textbf{4} & \textbf{-} & \textbf{99.08} & \textbf{97.66} \\
        & 0.05  & 0.05 & 25 & 5 & 4 & - & 98.63 & 97.87 \\
        & 0.05  & 0.05 & 30 & 5 & 4 & - & 98.32 & 97.94 \\
    \bottomrule
  \end{tabular}}
\end{table}

\begin{table}[htbp]
\caption{Results of all of algorithms on FMNIST with Setting 1.}
  \label{detail-result-dnn3-5}
  \centering
  \scalebox{0.85}{
  \begin{tabular}{ccccccccc}
    \toprule
    \multirow{2}{*}{Method} & \multirow{2}{*}{$\eta_1$} & \multirow{2}{*}{$\eta_2$} & \multirow{2}{*}{$\lambda$} & \multirow{2}{*}{$ K $}  & \multirow{2}{*}{$ E $} & \multirow{2}{*}{$\alpha $} &  \multicolumn{2}{c}{Acc.(\%)}\\
    \cmidrule(r){8-9}
        &   & & & & & &{PM}  &{GM}\\
    \midrule
        \multirow{1}{*}{Fedavg~\cite{mcmahan2017communication}}
        & 0.05  & - & - & 5 & - & - & - & 79.11 \\
    \midrule
        \multirow{1}{*}{HierFAVG~\cite{liu2020client}}
        & 0.05  & - & - & 5 & 4 & - & - & 83.55 \\
    \midrule
        \multirow{2}{*}{HeurFedAMP~\cite{huang2021personalized}}
        & 0.05  & 0.05 & - & 5 & - & 0.1 & 95.00 & -  \\
        & \textbf{0.05}  & \textbf{0.05}  & \textbf{-} & \textbf{5} & \textbf{-} & \textbf{0.5} & \textbf{98.28} & - \\
    \midrule
        \multirow{1}{*}{Fedprox~\cite{li2018federated}}
        & 0.05  & - & 0.001 & 5 & - & - & - & 78.67 \\
    \midrule
        \multirow{1}{*}{PerFedavg~\cite{fallah2020personalized}}
        & 0.05  & - & - & 5 & - & - & 96.95 & - \\
    \midrule
        \multirow{5}{*}{pFedMe~\cite{t2020personalized}}
        & 0.05  & 0.05 & 5 & 5 & - & - & {98.84} & {79.76} \\
        & 0.05  & 0.05 & 15 & 5 & - & - & {98.80} &{80.22} \\
        & 0.05  & 0.05 & 20 & 5 & - & - & {98.81} &{79.66} \\
        & {0.05}  & {0.05}  & {25} &{5} & {-} & {-} & {98.90} & {80.19} \\
        & \textbf{0.05}  & \textbf{0.05} & \textbf{30} & \textbf{5} & \textbf{-} & \textbf{-} & \textbf{98.94} & \textbf{80.57} \\
    \midrule
        \multirow{1}{*}{pFedGP~\cite{achituve2021personalized}}
        & 0.05 & - & - & 5 & - & - & 98.88 & - \\
    \midrule
        \multirow{5}{*}{\color{black}sFedHP (Ours)}
        & 0.05  & 0.05 & 5 & 5 & 4 & - & 98.55 & 84.01 \\
        & 0.05  & 0.05 & 15 & 5 & 4 & - & 97.61 & 82.75 \\
        & \textbf{0.05}  & \textbf{0.05}  & \textbf{20} & \textbf{5} & \textbf{4} & \textbf{-} & \textbf{98.42} & \textbf{85.64} \\
        & 0.05  & 0.05 & 25 & 5 & 4 & - & 95.86 & 86.51 \\
        & 0.05  & 0.05 & 30 & 5 & 4 & - & 91.94 & 86.86 \\
    \bottomrule
  \end{tabular}}
\end{table}

\begin{table}[htbp]
\caption{Results of all of algorithms on FMNIST with Setting 2.}
  \label{detail-result-dnn3-6}
  \centering
  \scalebox{0.85}{
  \begin{tabular}{ccccccccc}
    \toprule
    \multirow{2}{*}{Method} & \multirow{2}{*}{$\eta_1$} & \multirow{2}{*}{$\eta_2$} & \multirow{2}{*}{$\lambda$} & \multirow{2}{*}{$ K $}  & \multirow{2}{*}{$ E $} & \multirow{2}{*}{$\alpha $} &  \multicolumn{2}{c}{Acc.(\%)}\\
    \cmidrule(r){8-9}
        &   & & & & & &{PM}  &{GM}\\
    \midrule
        \multirow{1}{*}{Fedavg~\cite{mcmahan2017communication}}
        & 0.05  & - & - & 5 & - & - & - & 84.48 \\
    \midrule
        \multirow{1}{*}{HierFAVG~\cite{liu2020client}}
        & 0.05  & - & - & 5 & 4 & - & - & 85.48 \\
    \midrule
        \multirow{2}{*}{HeurFedAMP~\cite{huang2021personalized}}
        & 0.05  & 0.05 & - & 5 & - & 0.1 & 97.43 & - \\
        & \textbf{0.05}  & \textbf{0.05}  & \textbf{-} & \textbf{5} & \textbf{-} & \textbf{0.5} & \textbf{98.97} & - \\
    \midrule
        \multirow{1}{*}{Fedprox~\cite{li2018federated}}
        & 0.05  & - & 0.001 & 5 & - & - & - & 84.33 \\
    \midrule
        \multirow{1}{*}{PerFedavg~\cite{fallah2020personalized}}
        & 0.05  & - & - & 5 & - & - & 96.47 & - \\
    \midrule
        \multirow{5}{*}{pFedMe~\cite{t2020personalized}}
         & 0.05  & 0.05 & 5 & 5 & - & - & {99.12} &{82.21} \\
        & 0.05  & 0.05 & 15 & 5 & - & - &{99.14} & {82.18} \\
        & 0.05  & 0.05 & 20 & 5 & - & - & {99.14} &{84.42} \\
        & \textbf{0.05}  & \textbf{0.05}  & \textbf{25} & \textbf{5} & \textbf{-} & \textbf{-} & \textbf{99.17} & \textbf{85.14} \\
        & 0.05  & 0.05 & 30 & 5 & - & - & {99.16} & {84.74} \\
    \midrule
        \multirow{1}{*}{pFedGP~\cite{achituve2021personalized}}
        & 0.05 & - & - & 5 & - & - & 99.20 & - \\
    \midrule
        \multirow{5}{*}{\color{black}sFedHP (Ours)}
        & 0.05  & 0.05 & 5 & 5 & 4 & - & 98.93 & 85.73 \\
        & 0.05  & 0.05 & 15 & 5 & 4 & - & 98.5 & 83.7 \\
        & \textbf{0.05}  & \textbf{0.05}  & \textbf{20} & \textbf{5} & \textbf{4} & \textbf{-} & \textbf{98.98} & \textbf{89.29} \\
        & 0.05  & 0.05 & 25 & 5 & 4 & - & 96.73 & 89.47 \\
        & 0.05  & 0.05 & 30 & 5 & 4 & - & 93.35 & 89.05 \\
    \bottomrule
  \end{tabular}}
\end{table}

\begin{table}[htbp]
\caption{Results of all of algorithms on CIFAR-10 with Setting 1.}
  \label{detail-result-dnn3-3}
  \centering
  \scalebox{0.85}{
  \begin{tabular}{ccccccccc}
    \toprule
    \multirow{2}{*}{Method} & \multirow{2}{*}{$\eta_1$} & \multirow{2}{*}{$\eta_2$} & \multirow{2}{*}{$\lambda$} & \multirow{2}{*}{$ K $}  & \multirow{2}{*}{$ E $} & \multirow{2}{*}{$\alpha $} &  \multicolumn{2}{c}{Acc.(\%)}\\
    \cmidrule(r){8-9}
        &   & & & & & &{PM}  &{GM}\\
    \midrule
        \multirow{1}{*}{Fedavg~\cite{mcmahan2017communication}}
        & 0.05  & - & - & 5 & - & - & - & 42.36 \\
    \midrule
        \multirow{1}{*}{HierFAVG~\cite{liu2020client}}
        & 0.05  & - & - & 5 & 4 & - & - & 52.76 \\
    \midrule
        \multirow{2}{*}{HeurFedAMP~\cite{huang2021personalized}}
        & 0.05  & 0.05 & - & 5 & - & 0.1 & 71.36 & - \\
        & \textbf{0.05}  & \textbf{0.05}  & \textbf{-} & \textbf{5} & \textbf{-} & \textbf{0.5} & \textbf{82.26} & - \\
    \midrule
        \multirow{1}{*}{Fedprox~\cite{li2018federated}}
        & 0.05  & - & 0.001 & 5 & - & - & - & 39.10 \\
    \midrule
        \multirow{1}{*}{PerFedavg~\cite{fallah2020personalized}}
        & 0.05  & - & - & 5 & - & - & 77.20 & - \\
    \midrule
        \multirow{5}{*}{pFedMe~\cite{t2020personalized}}
        & 0.05  & 0.05 & 5 & 5 & - & - & {85.46} & {34.75} \\
        & 0.05  & 0.05 & 15 & 5 & - & - & {85.45} & {43.26} \\
        & 0.05  & 0.05 & 20 & 5 & - & - & {87.33} & {48.25} \\
        & 0.05  & 0.05 & 25 & 5 & - & - & {88.36} & {52.39} \\
        & \textbf{0.05}  & \textbf{0.05}  & \textbf{30} & \textbf{5} & \textbf{-} & \textbf{-} & \textbf{88.45} & \textbf{58.66} \\
    \midrule
        \multirow{1}{*}{pFedGP~\cite{achituve2021personalized}}
        & 0.05 & - & - & 5 & - & - & 85.14 & - \\
    \midrule
        \multirow{5}{*}{\color{black}sFedHP (Ours)}
        & 0.05  & 0.05 & 5 & 5 & 4 & - & {78.00} & {47.91} \\
        & 0.05  & 0.05 & 15 & 5 & 4 & - & {75.53} & {68.48} \\
        & \textbf{0.05}  & \textbf{0.05} & \textbf{20} & \textbf{5} & \textbf{4} & \textbf{-} & \textbf{83.95} & \textbf{78.44} \\
        & {0.05}  & {0.05}  & {25} & {5} &{4} & {-} &{76.98} &{65.83} \\
        & 0.05  & 0.05 & 30 & 5 & 4 & - & {76.24} & {66.39} \\
    \bottomrule
  \end{tabular}}
\end{table}

\begin{table}[htbp]
\caption{Results of all of algorithms on CIFAR-10 with Setting 2.}
  \label{detail-result-dnn3-4}
  \centering
  \scalebox{0.85}{
  \begin{tabular}{ccccccccc}
    \toprule
    \multirow{2}{*}{Method} & \multirow{2}{*}{$\eta_1$} & \multirow{2}{*}{$\eta_2$} & \multirow{2}{*}{$\lambda$} & \multirow{2}{*}{$ K $}  & \multirow{2}{*}{$ E $} & \multirow{2}{*}{$\alpha $} &  \multicolumn{2}{c}{Acc.(\%)}\\
    \cmidrule(r){8-9}
        &   & & & & & &{PM}  &{GM}\\
    \midrule
        \multirow{1}{*}{Fedavg~\cite{mcmahan2017communication}}
        & 0.05  & - & - & 5 & - & - & - & 81.74 \\
    \midrule
        \multirow{1}{*}{HierFAVG~\cite{liu2020client}}
        & 0.05  & - & - & 5 & 4 & - & - & 87.28 \\
    \midrule
        \multirow{2}{*}{HeurFedAMP~\cite{huang2021personalized}}
        & 0.05  & 0.05 & - & 5 & - & 0.1 & 83.51 & - \\
        & \textbf{0.05}  & \textbf{0.05}  & \textbf{-} & \textbf{5} & \textbf{-} & \textbf{0.5} & \textbf{88.13} & - \\
    \midrule
        \multirow{1}{*}{Fedprox~\cite{li2018federated}}
        & 0.05  & - & 0.001 & 5 & - & - & - & 72.98 \\
    \midrule
        \multirow{1}{*}{PerFedavg~\cite{fallah2020personalized}}
        & 0.05  & - & - & 5 & - & - & 84.92 & - \\
    \midrule
        \multirow{5}{*}{pFedMe~\cite{t2020personalized}}
        & 0.05  & 0.05 & 5 & 5 & - & - & {89.22} & {51.70} \\
        & 0.05  & 0.05 & 15 & 5 & - & - & {90.08} & {62.06} \\
        & 0.05  & 0.05 & 20 & 5 & - & - & {92.08} & {71.77} \\
        & {0.05}  & {0.05}  & {25} & {5} & {-} & {-} &{93.17} &{86.95} \\
        & \textbf{0.05}  & \textbf{0.05} & \textbf{30} & \textbf{5} & \textbf{-} & \textbf{-} & \textbf{93.40} & \textbf{90.31} \\
    \midrule
        \multirow{1}{*}{pFedGP~\cite{achituve2021personalized}}
        & 0.05 & - & - & 5 & - & - & 85.98 & - \\
    \midrule
        \multirow{5}{*}{\color{black}sFedHP (Ours)}
        & 0.05  & 0.05 & 5 & 5 & 4 & - & 88.44 & 83.17 \\
        & 0.05  & 0.05 & 15 & 5 & 4 & - & 88.86 & 87.62 \\
        & 0.05  & 0.05 & 20 & 5 & 4 & - & 92.39 & 92.42 \\
        & 0.05  & 0.05 & 25 & 5 & 4 & - & 93.16 & 93.27 \\
        & \textbf{0.05}  & \textbf{0.05}  & \textbf{30} & \textbf{5} & \textbf{4} & \textbf{-} & \textbf{93.98} & \textbf{94.09} \\
    \bottomrule
  \end{tabular}}
\end{table}

{\color{black}
Table~\ref{dif-alg-table} shows the global model performance of each algorithm. We maintain $|\mathcal{D}|=20$, $T=800$, $K=5$, and $\eta_1=\eta_2=0.05$ across all algorithms and fine-tune other fundamental hyperparameters. We found that hierarchical federated learning methods, such as sFedHP and HierFAVG, significantly outperformed other federated learning methods, including FedAvg, FedProx, and pFedMe, in the device-cloud structure on non-i.i.d. data. The proposed algorithm, \texttt{sFedHP}, outperforms the comparative algorithms concerning the global model in all settings by more than $1\%$ (Setting 1 on MNIST), $2\%$ (Setting 2 on MNIST), $2\%$ (Setting 1 on FMNIST), $3\%$ (Setting 2 on FMNIST), $19\%$ (Setting 1 on CIFAR-10), and $3\%$ (Setting 2 on CIFAR-10).

We present more detailed results for fine-tuning hyperparameters, including methods without a global model, in Tables~\ref{detail-result-dnn3-1} to~\ref{detail-result-dnn3-6}. Here, $E$ represents the number of edge servers in the hierarchical framework, and $\alpha$ represents the proportion of the client model that does not interact with the global model in HeurFedAMP. For HeurFedAMP, we tune $\alpha\in[0.1,0.5]$, where $\alpha$ represents the proportion of the client model that does not interact with the global model. For Fedprox, we tune $\lambda\in\{0.001,0.01,0.1,1\}$ following the setting in \cite{li2018federated}. For pFedMe and \texttt{sFedHP}, we use the same setting as in \cite{t2020personalized} with $\lambda\in\{5,15,20,25,30\}$. For pFedGP, we use the same setting as in \cite{achituve2021personalized} for other basic hyperparameters. We find that \texttt{sFedHP} outperforms all compared algorithms in the global model (GM) and achieves state-of-the-art results in the personalized model (PM).
}

\begin{figure}[!tp]
	\centering
	\subfloat{
	\includegraphics[width=1.65in]{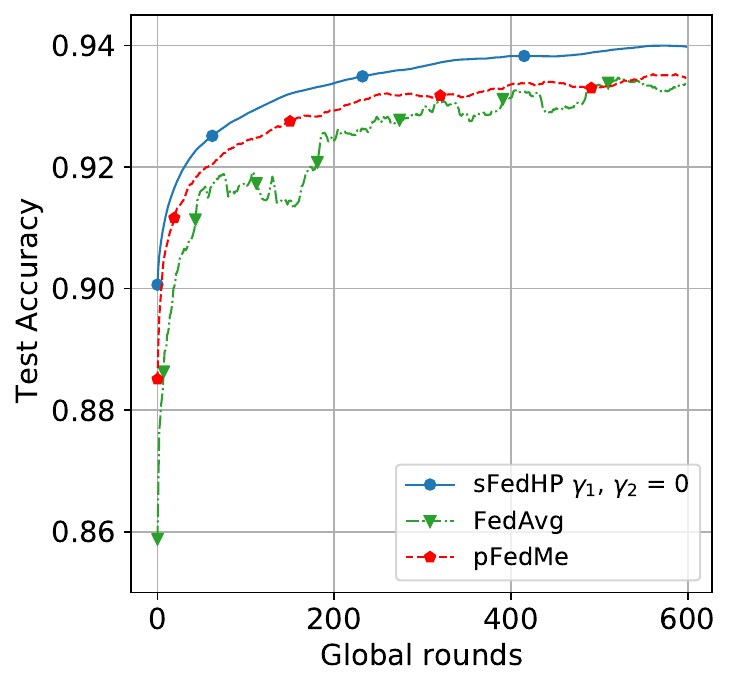}
	}
	\subfloat{
	\includegraphics[width=1.65in]{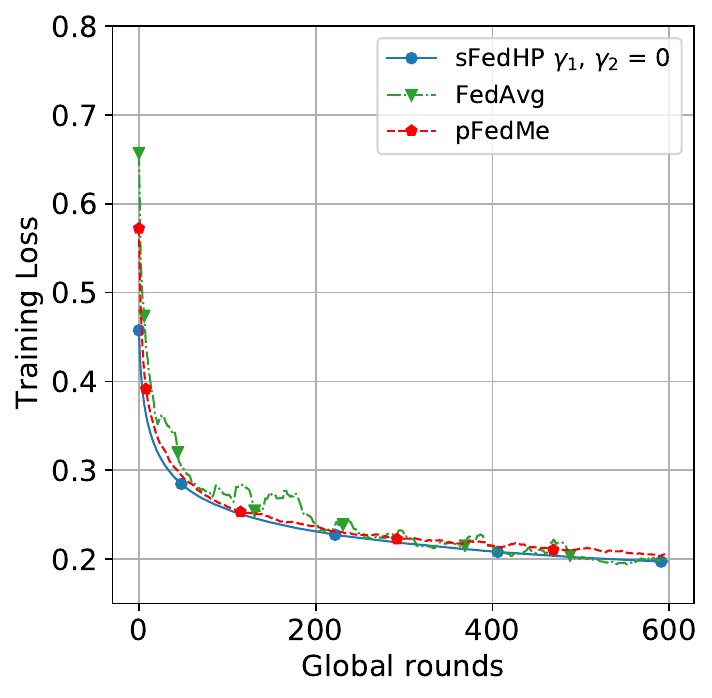}
	}
	\caption{$\mu$-Strongly Convex Performance comparison of \texttt{sFedHP}, FedAvg and pFedMe in MNIST ($\eta=0.05, R=20, |\mathcal{D}|=20, \beta=1$ for all experiments, $\lambda_1=\lambda_2=\lambda=25$ for \texttt{sFedHP} and pFedMe)}
	\label{fig_strong_convex}
\end{figure}

\subsection{More results}


We compare the global-model between \texttt{sFedHP}, pFedMe~\cite{t2020personalized} and hierarchical FedAvg~\cite{liu2020client} on $\mu$-strongly convex and nonconvex situactions. Similarly with ~\cite{t2020personalized}, in each round of local training, the client uses $K=5$ gradient-based iterations to obtain an approximated optimal local model, i.e., solve \eqref{H-min} in \texttt{sFedHP}. An $\ell_2$-regularized multinomial logistic regression model (MLR) is used for $\mu$-strong convex situation, while a deep neural netwrok (DNN) with two hidden layers of size [500, 200] is used for nonconvex situation. In our experiments, $\frac{3}{4}$ datasets are for training and the others are for testing. All experiments were conducted on a NVDIA Quadro RTX 6000 environment, and the code based on PyTorch is available online.

We use both sparsity and non-sparsity settings for \texttt{sFedHP}. In the sparsity setting, we set $\gamma_1=\gamma_2=0.001$ at the beginning, while set $\gamma_1$  and $\gamma_2$ to a tiny number respectively when the client model sparsity is lower than $0.2$ and after training 100 global rounds. Since according to our theoretical results, a tiny $\gamma_1, \gamma_2$ can reduce the convergence cost. In non-sparsity setting,  we set $\gamma_1=\gamma_2=0$.

Fig.~\ref{fig_strong_convex} shows the performance of \texttt{sFedHP}, hierarchical FedAvg and pFedMe in $\mu$-strongly convex situation. Since the MLR model is already lightweight enough, we set $\gamma_1=\gamma_2=0$ for \texttt{sFedHP}. In Fig.~\ref{fig_strong_convex}, we can see that both \texttt{sFedHP} and pFedMe perform better than hierarchical FedAvg, while \texttt{sFedHP} obtain higher test accuracy and convergence speed, which shows that the hierarchical personalization scheme is more suitable for solving statistical diversity problem.

\begin{figure*}[!tp]
	\centering
	\subfloat{
	\includegraphics[width=1.8in]{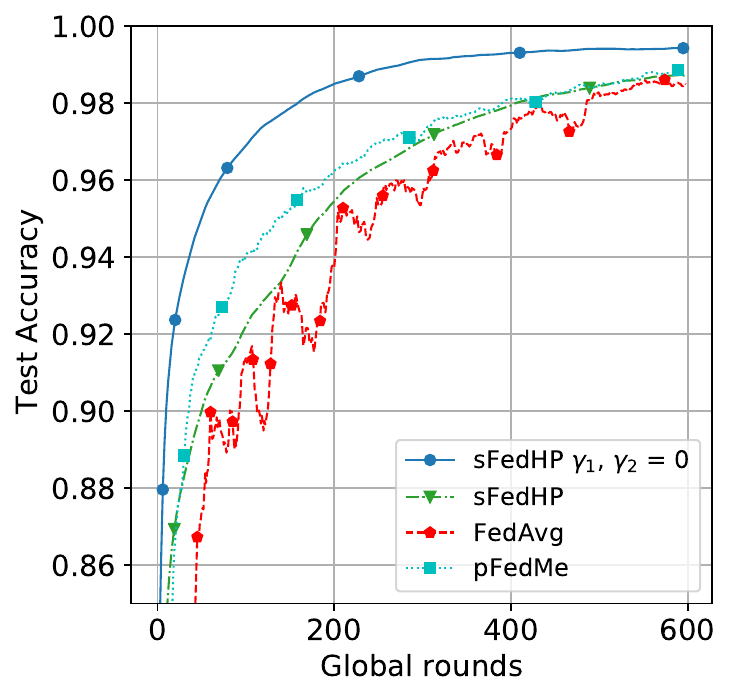}
	}
	\subfloat{
	\includegraphics[width=1.8in]{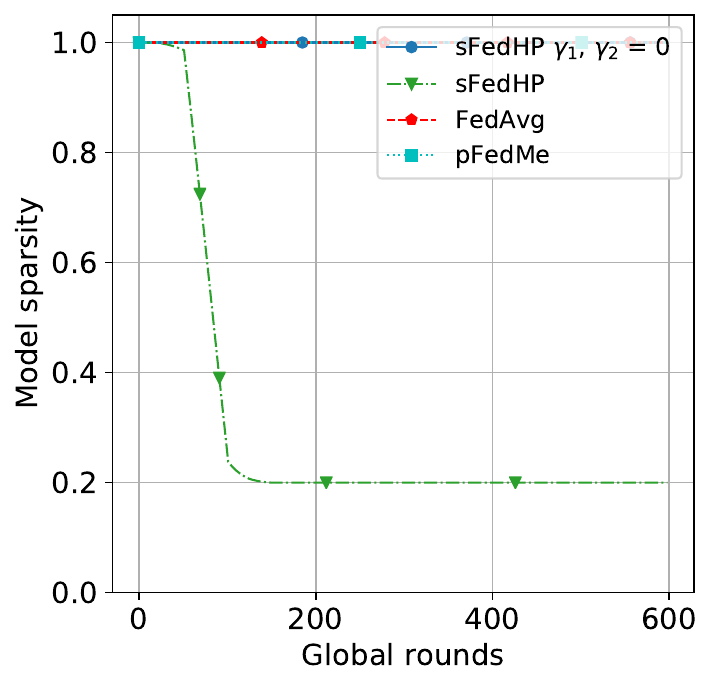}
	}
	\subfloat{
	\includegraphics[width=1.8in]{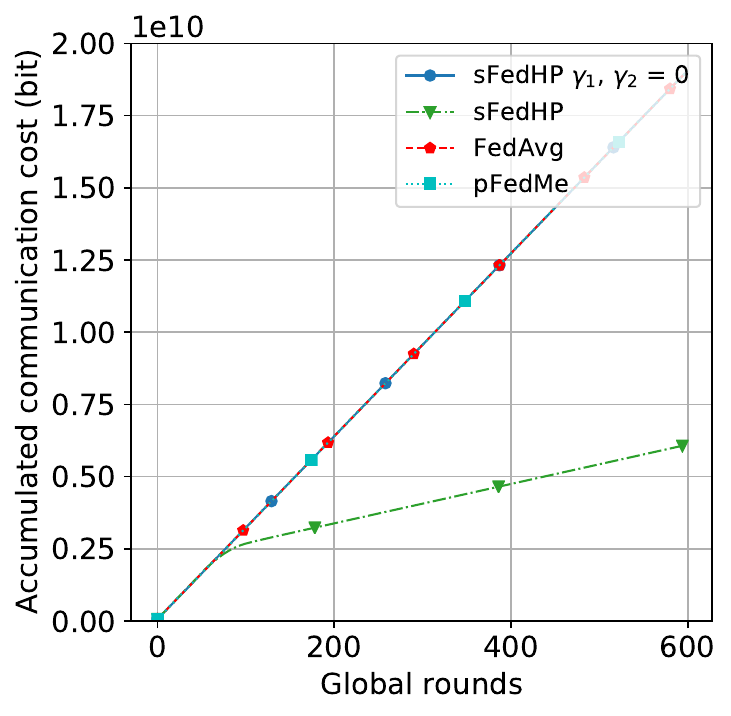}
	}
	\caption{Nonconvex Performance comparison of \texttt{sFedHP}, FedAvg and pFedMe in MNIST ($\eta=0.05, R=20, |\mathcal{D}|=20, \beta=1$ for all experiments, $\lambda_1=\lambda_2=\lambda=25$ for \texttt{sFedHP} and pFedMe)}
	\label{fig_dif_alg}
\end{figure*}

Fig.~\ref{fig_dif_alg} shows the performance of \texttt{sFedHP}, hierarchical FedAvg and pFedMe in nonconvex situation. We test sparsity and non-sparsity settings for \texttt{sFedHP} in accuracy, model sparsity and the accumulative communication. 
We set 20\% sparsity, the proportion of non-zero parameters, as a lower bound to preserve performance. 
According to Fig.~\ref{fig_dif_alg}, the proposed \texttt{sFedHP} can reduce the communication cost while achieve good test accuracy simultaneously. Specific, \texttt{sFedHP} obtain similar performance in a global model with pFedMe while reducing communication costs by 80\% and obtain higher performance when using a non-sparsity setting.

\begin{remark}
The accumulated communication cost in Fig.~\ref{fig_dif_alg} is calculated based on one client in the client-cloud FL or one edge server in the client-edge-cloud FL since the client and the edge server are nearby between which low-cost communication is possible.
Consider the DNN model with 79510 parameters,  whose non-zero parameters are quantized in 64 bits while zero parameters are quantized in 1 bit. 79510 bits location parameters are needed to mark the positions of zero parameters in \texttt{sFedHP}.
This section aims to demonstrate the excellent sparsity of sFedHP qualitatively. We focused on analyzing the ability of sFedHP to process non-i.i.d. data under different settings. More rigorous communication cost analysis needs to cooperate with the encoding and decoding process.
\end{remark}

In summary, \texttt{sFedHP} performs better that pFedMe and hierarchical FedAvg in test accuracy, convergence rate, and communication cost in $\mu$-strongly convex and nonconvex situations. Furthermore, the hierarchical personalization scheme is more suitable for solving statistical diversity problems.

\begin{figure}[!tp]
	\centering
	\subfloat{
	\includegraphics[width=1.65in]{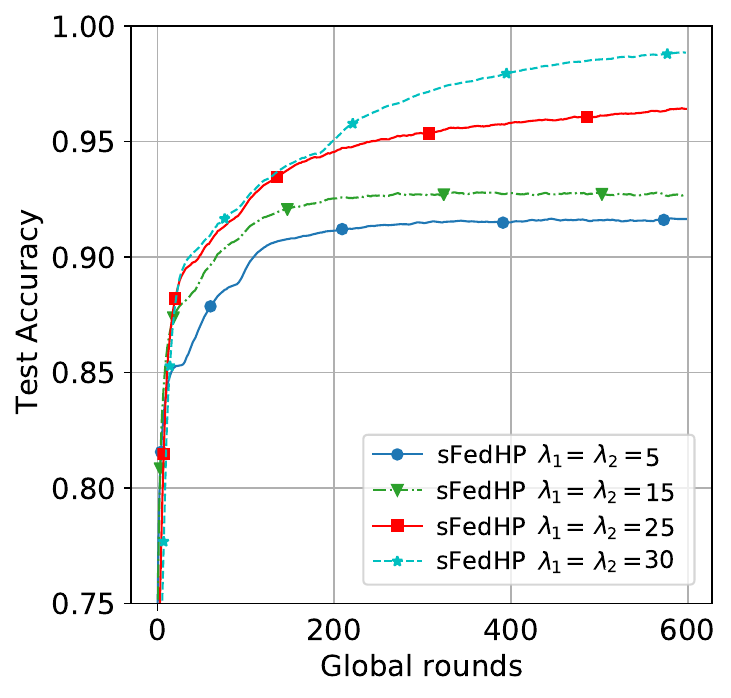}
	}
	\subfloat{
	\includegraphics[width=1.65in]{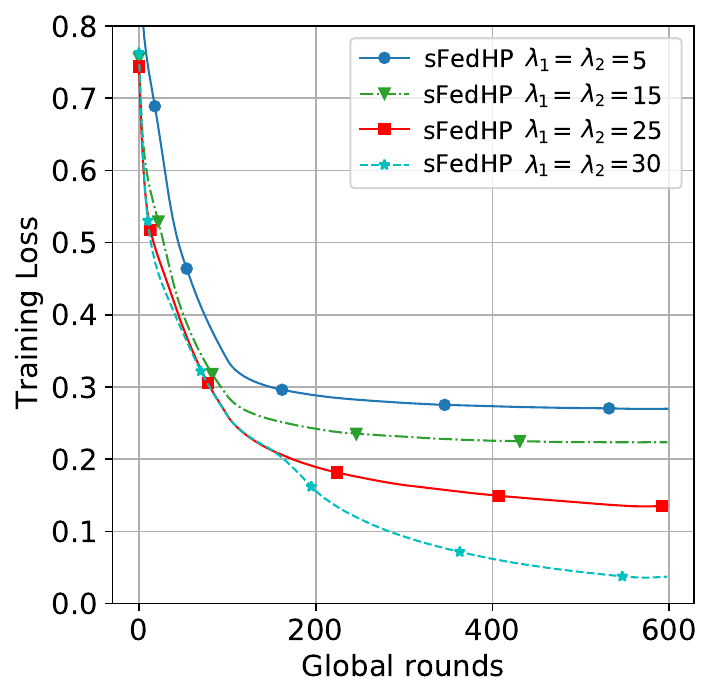}
	}
	\caption{\color{black}Effect of $\lambda$ on the convergence of \texttt{sFedHP} on MNIST ($\rho=6\times10^{-5}, \gamma_1=\gamma_2=0.008, R=20, K=5, \beta=1$).}
	\label{fig_change_lambda}
\end{figure}

\begin{figure}[!tp]
	\centering
	\subfloat{
	\includegraphics[width=1.65in]{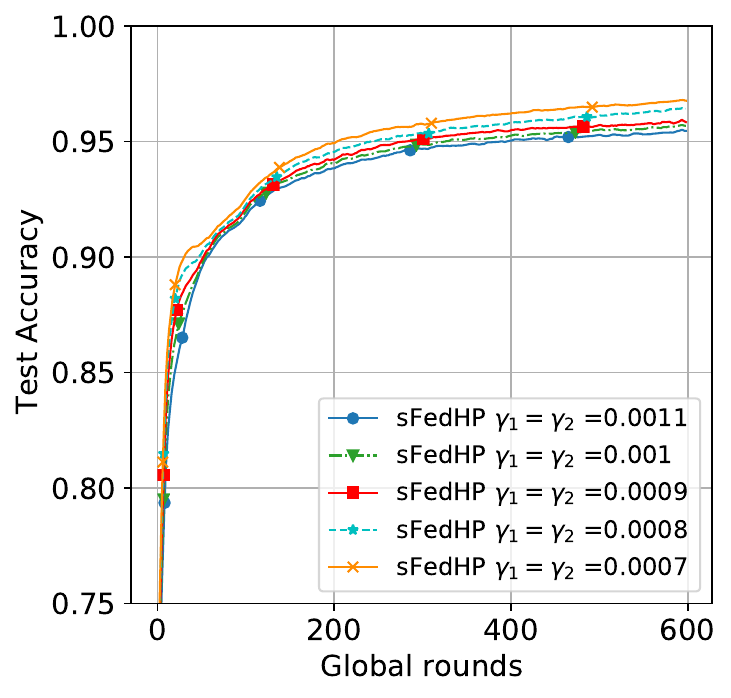}
	}
	\subfloat{
	\includegraphics[width=1.65in]{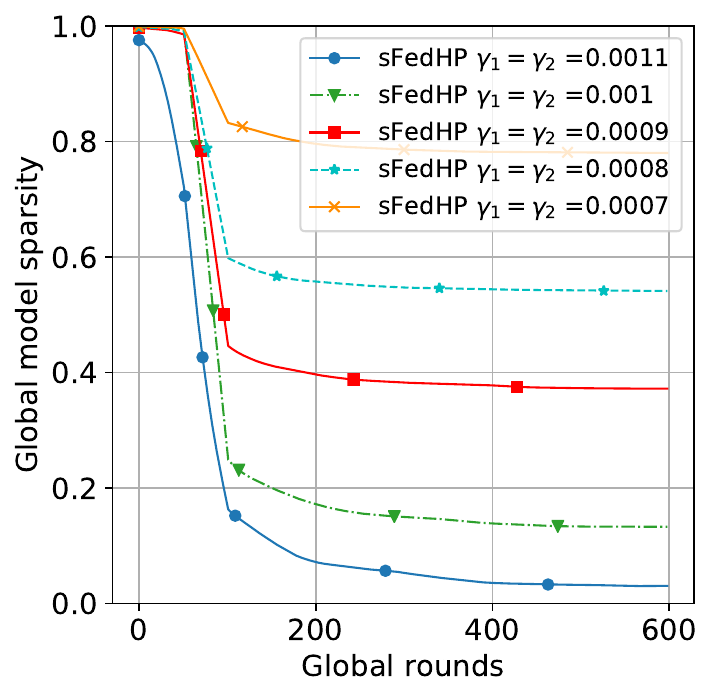}
	}
	\caption{Effect of $\gamma$ on the convergence of \texttt{sFedHP} on MNIST ($\rho=6\times10^{-5}, R=20, \lambda_1=\lambda_2=25, K=5, \beta=1$).}
	\label{fig_change_gamma}
\end{figure}

\subsection{Effect of hyperparameters}\label{sec_result_hyper}

We empirically study the effect of different hyperparameters in {\texttt{sFedHP}}.

\textbf{Effects of  $\lambda$:} According to Fig.~\ref{fig_change_lambda}, properly increasing $\lambda$ can effectively improve the test accuracy and convergence rate for \texttt{sFedHP}. And we find that an oversize $\lambda_1$ and $\lambda_2$ may cause gradient explosion.

\textbf{Effects of $\gamma$:} Fig.~\ref{fig_change_gamma} shows the relationship of $\gamma$ with the model sparsity and the convergence speed, we can see that increasing $\gamma$ will reduce the communication speed and global accuracy.

\section{Conclusion} \label{sec_conclusion}
{\color{black}
In this paper, we propose \texttt{sFedHP} as a sparse hierarchical personalized FL algorithm that can greatly remiss the statistical diversity issue to improve the FL performance and reduce the communication cost in FL. 
Our approach uses an approximated $\ell_1$-norm and the hierarchical proximal mapping to generate the loss function.
The hierarchical proximal mapping enables the personalized edge-model, and client-model optimization can be decomposed from the global-model learning, which allows \texttt{sFedHP} parallelly optimizes the personalized edge-model and client-model by solving a tri-level problem. Theoretical results present that the sparse constraint in \texttt{sFedHP} only reduces the convergence speed to a small extent. Experimental results further demonstrate that \texttt{sFedHP} outperforms the client-edge-cloud hierarchical FedAvg and many other personalized FL methods based on local customization under different settings.
}

\appendix[Proof of the Results]


\subsection*{Proof of Lemma~\ref{lemma_3}}

\begin{proof}
First, we have
\begin{align}
\label{Ewt}
&~{\mathbb{E}}\left[\left\|\bm w^{t+1}-\bm w^{*}\right\|^{2}\right]={\mathbb{E}}\left[\left\|\bm w^{t}-\tilde{\eta} g^{t}-\bm w^{*}\right\|_2^{2}\right]  \\
=&~{\mathbb{E}}\left[\left\|\bm w^{t}-\bm w^{*}\right\|_2^{2}\right]-2 \tilde{\eta} {\mathbb{E}}\left[\left\langle g^{t}, \bm w^{t}-\bm w^{*}\right\rangle\right]+\tilde{\eta}^{2} \mathbb{E}\left[\left\|g^{t}\right\|_2^{2}\right] \nonumber
\end{align}
and the second term of \eqref{Ewt} is as follow
\begin{align}
&~~~~-{\mathbb{E}}\left\langle g^{t}, \bm w^{T}-\bm w^{*}\right\rangle \notag\\
&=-\frac{1}{N R} \sum_{i, r}^{N, R}(\left\langle g_{i}^{t,r}-\nabla F_{i}\left(\bm w^{t}\right), \bm w^{t}-\bm w^{*}\right\rangle\notag\\
&~~~~~~~~~~~~~~~~~~~~~~~~~+\left\langle\nabla F_{i}\left(w^{t}\right), \bm w^{t}-\bm w^{*}\right\rangle)\notag\\
&\overset{(a)}{\leq}\frac{1}{2N R} \sum_{i, r}^{N, R}\left(\frac{\left\|g_{i}^{t,r}-\nabla F_{i}\left(\bm w^{t}\right)\right\|_2^{2}}{\mu_{F}}+{\mu_{F}}\left\|\bm w^{t}-\bm w^{*}\right\|_2^{2}\right)\notag\\
&~~~~~~~~~~~~~~~~~~~~~~~~~-\frac{1}{NJ} \sum_{i,j=1}^{NJ}\left(\left\langle\nabla F_{i,j}\left(\bm w^{t}\right), \bm w^{t}-\bm w^{*}\right\rangle\right) \notag\\
&\overset{(b)}{\leq}\frac{1}{2N R} \sum_{i, r}^{N, R}\left(\frac{\left\|g_{i}^{t,r}-\nabla F_{i}\left(\bm w^{t}\right)\right\|_2^{2}}{\mu_{F}}+{\mu_{F}}\left\|\bm w^{t}-\bm w^{*}\right\|_2^{2}\right)\notag\\
&~~~~~~~~~~~~~~~~~~~~~~~+\left(F\left(\bm w^{*}\right)-F\left(\bm w^{t}\right)-\frac{\mu_{F}}{2}\left\|\bm w^{t}-\bm w^{*}\right\|_2^{2} \right)\notag\\
&=\frac{1}{2N R} \sum_{i, r}^{N, R}\left(\frac{\left\|g_{i}^{t,r}-\nabla F_{i}\left(\bm w^{t}\right)\right\|_2^{2}}{\mu_{F}}\right)+F\left(\bm w^{*}\right)-F\left(\bm w^{t}\right).\label{eq-la3-2}
\end{align}
where $(a)$ follows by the Peter Paul inequality and the Jensen
s inequality and $(b)$ is due to Theorem~\ref{theorem_1}.

From equations $(18)$ and $(19)$ in \cite{t2020personalized}, we have
\begin{align}
    \left\|g^{t}\right\|^{2}&\leq \frac{3}{N R} \sum_{i, r}^{N, R}\left\|g_{i}^{t,r}-\nabla F_{i}\left(\bm w^{t}\right)\right\|_2^{2}\notag\\
    &~~~~~~+3 \left\|\frac{1}{S} \sum_{i \in \mathcal{S}^{t}} \nabla F_{i}\left(\bm w^{t}\right)-\nabla F\left(\bm w^{t}\right)\right\|_2^{2}\notag\\
    &~~~~~~+6 L_{F}\left(F\left(\bm w^{t}\right)-F\left(\bm w^{*}\right)\right).\label{eq-la3-3}
\end{align}

Taking expectation of the second term of \eqref{eq-la3-3} w.r.t edge server sampling, we have
\begin{align}
	&~\mathbb{E}_{S_t}\left\|\frac{1}{S} \sum_{i \in \mathcal{S}^{t}} \nabla F_{i}\left(\bm w^{t}\right)-\nabla F\left(\bm w^{t}\right)\right\|_2^{2}\notag\\
	\overset{(a)}{\leq}&~
	\frac{1}{SJ}\sum_{j=1}^{J}\mathbb{E}_{S_t}\left(\sum_{i \in \mathcal{S}^{t}} \left\|\nabla F_{i,j}\left(\bm w^{t}\right)-\nabla F\left(\bm w^{t}\right)\right\|_2^{2}\right)\notag\\
	\overset{(b)}{=}&~
	\frac{1}{SJ}\sum_{i,j}^{NJ}\left( \left\|\nabla F_{i,j}\left(\bm w^{t}\right)-\nabla F\left(\bm w^{t}\right)\right\|_2^{2}\right)\mathbb{E}_{S_t}[\mathbb{I}_{i\in S_t}]\notag\\
	{\leq}&~
	\frac{1}{NJ}\sum_{i,j}^{NJ}\left( \left\|\nabla F_{i,j}\left(\bm w^{t}\right)-\nabla F\left(\bm w^{t}\right)\right\|_2^{2}\right)\label{eq-la3-4}
\end{align}
where $(a)$ is due to the Jensen's inequality; $\mathbb{I}_{X}$ in $(b)$ is an indicator function of an event $X$; and $\mathbb{E}_{S_t}[\mathbb{I}_{i\in S_t}]=\frac{S}{N}$.

By substituting \eqref{eq-la3-2}, \eqref{eq-la3-3}, \eqref{eq-la3-4} into \eqref{Ewt}, we finish the proof of Lemma~\ref{lemma_3}.
\end{proof}

\subsection*{Proof of Lemma~\ref{lemma_1}}
\begin{proof}
By noting that the last term of \eqref{H-min} only w.r.t. $\bm y_{i,j}^{t,r}$ we have
\begin{align}
     \bm{\tilde \theta}_{i,j}(\!\bm y_{i,j}^{t,r}\!) \!&=\! \!\arg\min_{\boldsymbol{\theta}_{i,j} \in \mathbb{R}^{d}}\!  \tilde \ell_{i,j}\left(\!\boldsymbol{\theta}_{i,j}\!\right) \!+\! \frac{\lambda_1}{2} \! \left\|\!\bm \theta_{i,j}-\bm y_{i,j}^{t,r}\!\right\|_2^{2} \!+\! \gamma_1\phi_\rho(\!\bm \theta_{i,j}\!)\notag.\\
     \bm{\hat \theta}_{i,j}(\!\bm y_{i,j}^{t,r}\!) \!&= \!\arg\min _{\boldsymbol{\theta}_{i,j} \in \mathbb{R}^{d}}\!	 \hat\ell_{i,j}\left(\!\boldsymbol{\theta}_{i,j}\!\right)\!+\!\frac{\lambda_1}{2}\!\left\|\!\bm \theta_{i,j}-\bm y_{i,j}^{t,r}\!\right\|_2^{2} \!+\! \gamma_1\phi_\rho(\!\bm \theta_{i,j}\!)\notag.
\end{align}

Denote $h_{i,j}\left(\bm \theta_{i,j};\bm y_{i,j}^{t,r}\right)=\ell_{i,j}\left(\boldsymbol{\theta}_{i,j}\right)+\gamma_1\phi_\rho(\bm \theta_{i,j})+\frac{\lambda_1}{2}\left\|\bm \theta_{i,j}-\bm y_{i,j}^{t,r}\right\|_2^{2}$. Then, $h_i\left(\bm \theta_{i,j};\bm y_{i,j}^{t,r}\right)$ is $(\lambda_1+\mu)$-strongly convex when $A.\ref{assumption_1}(a)$ holds and $(\lambda_1-L-\frac{\gamma_1}{\rho})$-strongly convex when $A.\ref{assumption_1}(b)$ and $\lambda_1>L+\frac{\gamma_1}{\rho}$ holds. Follow the proof in \cite{t2020personalized}, we obtain the result in Lemma~\ref{lemma_1}.
\end{proof}

\subsection*{Proof of Lemma~\ref{lemma_4}}

\begin{proof}
\begin{align}
    &~~~~\mathbb{E}\left[\left\|g_{i}^{t,r}-\nabla F_{i}\left(\bm w^{t}\right)\right\|_2^{2}\right] \notag\\
    &\overset{(a)}{\leq} 2 \mathbb{E}\left[\left\|g_{i}^{t,r}\!-\!\nabla F_{i}\left(\bm w_{i}^{t,r}\right)\right\|_2^{2}+\left\|\nabla F_{i}\left(\bm w_{i}^{t,r}\right)\!-\!\nabla F_{i}\left(\bm w^{t}\right)\right\|_2^{2}\right]\notag\\
    &\overset{(b)}
    {=}
    \frac{2}{J}\sum_{j=1}^{J}(\bar\lambda^2 \mathbb{E}\left[\left\|\tilde{\bm \theta}_{i,j}^{t,r+1}\!-\!\hat{\bm \theta}_{i,j}^{t,r+1}\right\|_2^{2}\right]+L_{F}^{2} \mathbb{E}\left[\left\|\bm w_{i}^{t,r}\!-\!\bm w^{t}\right\|_2^{2}\right])\notag\\
    & \leq 2\left(\bar\lambda^{2} \delta^{2}+L_{F}^{2} \mathbb{E}\left[\left\|\bm w_{i}^{t,r}-\bm w^{t}\right\|_2^{2}\right]\right)
\end{align}
where $(a)$ follows by Jensen's inequality; $(b)$ is due to Theorem~\ref{theorem_1} and  the first-order conditions $\lambda_1(\hat{\bm y}_{i,j}-\hat{\bm \theta}_{i,j})+\lambda_2(\hat{\bm y}_{i,j}-\bm w)=0$ from \eqref{sFedPer-2} with $\bar \lambda=\frac{\lambda_1\lambda_2}{\lambda_1+\lambda_2}$. Moreover, the last term is bounded by
\begin{align}
    &~\mathbb{E}\left[\left\|\bm w_{i}^{t,r}-\bm w^{t}\right\|_2^{2}\right]\notag\\
    \overset{(a)}
    \leq&~
    \frac{8 \tilde{\eta}}{\beta L_F} \left(3 \mathbb{E}\frac{1}{J}\sum_{j=1}^{J}\left[\left\|\nabla F_{i,j}\left(\bm w^{t}\right)\right\|_2^{2}\right]+\frac{2 \bar\lambda^{2} \delta^{2}}{R}\right)\notag
\end{align}
where $(a)$ follows by our proof of Lemma~2 in~\cite{liu2021personalized}. Thus we finish the proof of Lemma~\ref{lemma_4}.
\end{proof}

\subsection*{Proof of Lemma~\ref{lemma_2}}

\begin{proof}
We first prove case (a), we have
\begin{align}
    &~\frac{1}{NJ} \sum_{i=1,j=1}^{N,J}\left\|\nabla F_{i,j}(\bm w)-\nabla F(\bm w)\right\|_2^{2} \notag\\
    \overset{(a)}{=}&~
    \frac{1}{NJ} \sum_{i,j=1}^{N,j}\left\|\nabla F_{i,j}(\bm w)\right\|_2^{2}-\left\|\frac{1}{NJ} \sum_{p,q=1}^{N}\nabla F_{i,j}(\bm w)\right\|_2^{2}\notag\\
    \overset{(b)}{\leq}&~
    \frac{2}{NJ}\sum_{i,j=1}^{N,J}{\left\|\nabla F_{i,j}(\bm w)-\nabla F_{i,j}(\bm w^*)\right\|_2^{2}+\left\|\nabla F_{i,j}(\bm w^*)\right\|_2^{2}}\notag\\
    \overset{(c)}{\leq}&~
    4 L_{F}\left(F(\bm w)-F\left(\bm w^{*}\right)\right)+ \frac{2}{NJ} \sum_{i,j=1}^{N}\left\|\nabla F_{i,j}\left(\bm w^{*}\right)\right\|_2^{2}\notag
\end{align}
where $(a)$ follows by $\mathbb{E}\left[\|\bm x-\mathbb{E}\left[\bm x\right]\|_2^2\right]=\mathbb{E}\left[\|\bm x\|_2^2\right]-\mathbb{E}\left[\|\bm x\|_2\right]^2$; $(b)$ follows by the Jensen's inequality; $(c)$ follows by Theorem~\ref{theorem_1}, thus $F_i$ is $\mu_F$-strong convex and $L_F$-smooth with $\nabla F(\bm w^*)=0$.\\

We next prove case (b). For the minimization problem in \eqref{sFedPer-2}, we have the first-order conditions $\lambda_1(\hat{\bm y}_{i,j}-\hat{\bm \theta}_{i,j})+\lambda_2(\hat{\bm y}_{i,j}-\bm w)=0$ and $\nabla\ell_{i,j}(\hat{\bm\theta}_{i,j})-\lambda_1(\hat{\bm y}_{i,j}-\hat{\bm \theta}_{i,j})+\gamma_1\nabla\phi_\rho(\hat{\bm\theta}_{i,j})=0$, then we have
\begin{align}
\label{first_order}
	\nabla F_{i,j}(\bm w) &= \lambda_2(\bm w-\hat{\bm y}_{i,j})+\gamma_2\nabla\phi_\rho(\bm w)\notag\\
	&=\lambda_1(\hat{\bm y}_{i,j}-\hat{\bm \theta}_{i,j})+\gamma_2\nabla\phi_\rho(\bm w)\notag\\
	&=\nabla \ell_{i,j}(\hat{\bm \theta}_{i,j})+\gamma_1\nabla\phi_\rho(\hat{\bm \theta}_{i,j})+\gamma_2\nabla\phi_\rho(\bm w)\notag\\
	&=\nabla L_{i,j}(\hat{\bm \theta}_{i,j})+\gamma_2\nabla\phi_\rho(\bm w),
\end{align}
where  $L_{i,j}(\cdot)=\ell_{i,j}(\cdot)+\gamma_1\phi_\rho(\cdot)$.


Hence, we have
\begin{align}
    &~\left\|\nabla F_{i,j}(\bm w)-\nabla F(\bm w)\right\|_2^{2}\notag\\
    \overset{(a)}
    {\leq}&~ 2\left\|\nabla L_{i,j}\left(\hat{\bm\theta}_{i,j}\right)-\frac{1}{NJ} \sum_{p,q=1}^{NJ} \nabla L_{p,q}\left(\hat{\bm\theta}_{i,j}\right)\right\|_2^{2}\notag\\
	&~~~+2\left\|\frac{1}{NJ} \sum_{p,q=1}^{NJ}\left(\nabla L_{p,q}\left(\hat{\bm\theta}_{i,j}\right)-\nabla L_{p,q}\left(\hat{\bm\theta}_{p,q}\right)\right)\right\|_2^2, \notag
\end{align}
where $(a)$ follows by \eqref{first_order} and the Jensen's inequality. Taking the average over the number of clients, we have
\begin{align}
\label{bound-1}
    &~\frac{1}{NJ} \sum_{i,j=1}^{N,J}\left\|\nabla F_{i,j}(\bm w)-\nabla F(\bm w)\right\|_2^{2}\notag\\
	\overset{(a)}{\leq}&~
	2\sigma_{\ell}^2+\frac{2}{N^2J^2} \sum_{i,j=1}^{N,J}\sum_{p,q=1}^{N,J}\left\|\nabla L_{p,q}\left(\hat{\bm\theta}_{i,j}\right)\!-\!\nabla L_{p,q}\left(\hat{\bm\theta}_{p,q}\right)\right\|_2^2\notag\\
	\overset{(b)}{\leq}&~
	2\sigma_{\ell}^2+\frac{2(L+\frac{\gamma_1}{\rho})^2}{N^2J^2} \sum_{i,j=1}^{N,J}\sum_{p,q=1}^{N,J}\left\|\hat{\bm\theta}_{i,j}-\hat{\bm\theta}_{p,q}\right\|_2^2 ,
\end{align}
where $(a)$ follows by Assumption~\ref{assumption_3} and the Jensen's inequality; $(b)$ is due to the $(L+\frac{\gamma_1}{\rho})$-smoothness of $L_{i,j}(\cdot)$; and the last term
\begin{align}
\label{bound-2}
	&~\sum_{i,j=1}^{N,J}\sum_{p,q=1}^{N,J}\left\|\hat{\bm\theta}_{i,j}-\hat{\bm\theta}_{p,q}\right\|_2^2\notag\\
	\overset{(a)}{\leq}&~4\sum_{i,j=1}^{N,J}\sum_{p,q=1}^{N,J} \left (\left\|\hat{\bm\theta}_{i,j}-\hat{\bm y}_{i,j}-\frac{\gamma_2}{\lambda_1}\nabla\phi_\rho(\bm w)\right\|_2^2 \right. \notag\\
	&~~~~~~~~~~~~~~+\left\|\hat{\bm\theta}_{p,q}-\hat{\bm y}_{p,q}-\frac{\gamma_2}{\lambda_1}\nabla\phi_\rho(\bm w)\right\|_2^2 \notag\\
	&~~~~~~~~~~~~~~+\left\|\hat{\bm y}_{i,j}-\bm w-\frac{\gamma_2}{\lambda_2}\nabla\phi_\rho(\bm w)\right\|_2^2 \notag\\
	&~~~~~~~~~~~~~~\left. +\left\|\hat{\bm y}_{p,q}-\bm w-\frac{\gamma_2}{\lambda_2}\nabla\phi_\rho(\bm w)\right\|_2^2 \right)\notag\\
	\overset{(b)}
    {=}&~\frac{8NJ(\lambda_1^2+\lambda_2^2)}{\lambda_1^2\lambda_2^2}\sum_{i,j=1}^{N,J}\left\|\nabla F_{i,j}(\bm w)\right\|_2^2 ,
\end{align}
where $(a)$ is due to the Jensen's inequality; $(b)$ follows by \eqref{first_order} and re-arranging the terms. Substituting \eqref{bound-2} back to \eqref{bound-1} we have
\begin{align}
    &~\frac{1}{NJ} \sum_{i,j=1}^{N,J}\left\|\nabla F_{i,j}(\bm w)-\nabla F(\bm w)\right\|_2^{2}\notag\\
	\overset{(a)}{\leq}&~2\sigma_{\ell}^2\notag+\frac{16(L+\frac{\gamma_1}{\rho})^2(\lambda_1^2+\lambda_2^2)}{\lambda_1^2\lambda_2^2} \Bigg( \left\|\nabla F(\bm w)\right\|_2^2  \notag\\
	&~~~~~~~~~~~~~  +\frac{1}{NJ}\sum_{i,j=1}^{N,J}\left\|\nabla F_{i,j}(\bm w)-\nabla F(\bm w)\right\|_2^2 \Bigg)\notag\\
	\overset{(b)}{\leq}&~
	\frac{16(L+\frac{\gamma_1}{\rho})^2}{\lambda^2-16(L+\frac{\gamma_1}{\rho})^2}\left\|\nabla F(\bm w)\right\|_2^2+\frac{2\lambda^2}{\lambda^2-16(L+\frac{\gamma_1}{\rho})^2}\sigma_{\ell}^2,\notag
\end{align}
where $(a)$ follows by the equation $\mathbb{E}\left[\|\bm x-\mathbb{E}\left[\bm x\right]\|_2^2\right]=\mathbb{E}\left[\|\bm x\|_2^2\right]-\mathbb{E}\left[\|\bm x\|_2\right]^2$; $(b)$ is by re-arranging the terms with setting $\lambda = \frac{\lambda_1\lambda_2}{\sqrt{\lambda_1^2+\lambda_2^2}}$.
\end{proof}

\subsection*{Proof of Theorem~\ref{theorem_2}}

\subsubsection*{Completing the proof of Theorem~1}
\begin{proof}
Let $\tilde\eta \leq \min\left \{ \frac{1}{\mu_F}, \frac{1}{3L_F(3+128L_F/(\mu_F\beta))}\right \}$ hold, combining Lemma~\ref{lemma_4} and Lemma~\ref{lemma_2} with Lemma~\ref{lemma_3} yields
\begin{align}
&~\mathbb{E}\left[\|\bm w^{t+1}-\bm w^*\|_2^2\right] \notag\\
\overset{(a)}{\leq}&~ \Delta_t -\tilde\eta \mathbb{E}\left[ F_{}\left(\bm w^{t}\right)- F_{}\left(\bm w^{*}\right)\right]
+\tilde\eta M_1 +\tilde\eta^2 M_2, \notag
\end{align}
where  $\Delta_t$ denotes $\mathbb{E}\left[\|\bm w^{t}-\bm w^*\|_2^2\right]$,
$M=2-\tilde\eta L_F(18+\frac{768L_F}{\mu_F\beta})$,  $M_1=\frac{8\delta^2\bar\lambda^2}{\mu_F}$, $M_2=6(1+64\frac{L_F}{\mu_F\beta})\sigma_{F,1}^2 + \frac{128L_F\delta^2\bar\lambda^2}{\mu_F\beta}$, and $(a)$ is due to $\tilde\eta \leq \frac{1}{3L_F(3+128L_F/(\mu_F\beta))}$, then
\begin{align}
\frac{1}{T}\sum_{t=0}^{T-1} \mathbb{E}\left[ F_{}\left(\bm w^{t}\right)- F_{}\left(\bm w^{*}\right)\right]
\leq \frac{\Delta_0}{\tilde\eta T}+ M_1 +\tilde\eta M_2. \label{eq-t2-a}
\end{align}
Then, the proof of part (a) is finished.

We next prove the part (b), we have
\begin{align}
    &~\mathbb{E}\left[\left\|\tilde{\theta}_{i,j}^{T}-\bm w^{*}\right\|_2^{2}\right]\notag\\
    \overset{(a)}{\leq}&~ 5 \mathbb{E}(\left\|\tilde{\theta}_{i,j}^{T}-\hat{\theta}_{i,j}^{T}\right\|_2^{2}+\left\|\hat{\theta}_{i,j}^{T}-\hat{\bm y}_{i,j}^{T}-\frac{\gamma_2}{\lambda_1}\nabla\phi_{\rho}(\boldsymbol{\bm w^T})\right\|_2^{2}\notag\\
    &~~~~~~+\left\|\hat{\bm y}_{i,j}^{T}-\bm w^T-\frac{\gamma_2}{\lambda_2}\nabla\phi_{\rho}(\boldsymbol{\bm w^T})\right\|_2^{2}+\left\|\bm w^{T}-\bm w^{*}\right\|_2^{2}\notag\\
    &~~~~~~+\left\|\frac{\gamma_2(\lambda_1+\lambda_2)}{\lambda_1\lambda_2}\nabla\phi_{\rho}(\boldsymbol{\bm w^T})\right\|_2^2)\notag\\
    \overset{(b)}
    {\leq}&~ 5(\delta^{2}+\frac{\gamma_2^2}{\bar\lambda^2}\mathbb{E}\left[\left\|\nabla\phi_{\rho}(\boldsymbol{\bm w^T})\right\|_2^2\right]+\mathbb{E}\left[\left\|\bm w^{T}-\bm w^{*}\right\|_2^{2}\right])\notag\\
    &+\frac{10}{\lambda^2} \mathbb{E}\left[\left\|\nabla F_{i,j}\left(\bm w^{T}\right)-\nabla F_{i,j}\left(\bm w^{*}\right)\right\|_2^{2}+\left\|\nabla F_{i,j}\left(\bm w^{*}\right)\right\|_2^{2}\right]\notag\\
    \overset{(c)}{\leq}&~ 5(\delta^{2}+\frac{\gamma_2^2}{\bar\lambda^2}\mathbb{E}\left[\left\|\nabla\phi_{\rho}(\boldsymbol{\bm w^T})\right\|_2^2\right]+\mathbb{E}\left[\left\|\bm w^{T}-\bm w^{*}\right\|_2^{2}\right])\notag\\
    &+\frac{10}{\lambda^2} \mathbb{E}\left[L_F\left\|\bm w^{T}-\bm w^{*}\right\|_2^{2}+\left\|\nabla F_{i,j}\left(\bm w^{*}\right)\right\|_2^{2}\right],
\label{eq_theta-w}
\end{align}
where $(a)$ and $(b)$ are due to the Jensen's inequality and \eqref{first_order}; and $(c)$ is due to Theorem~\ref{theorem_1}. In addition, the second term of \eqref{eq_theta-w} is bounded by
\begin{align}
    \mathbb{E}\!\left[\!\left\|\!\nabla\phi_{\rho}({\bm{w}^T})\!\right\|_2^2\!\right]
    \!=\!\mathbb{E}\!\left[\!\sum_{i=1}^{d}\!\left(\!\frac{\sinh \left(\!\left|w^{T}_n\right| \!/\! \rho\!\right)}{\cosh \left(\!\left|w^{T}_n\right| \!/\! \rho\!\right)} \!\frac{w^{T}_n}{\left|w^{T}_n\right|}\!\right)^2\!\right]
    \!\leq \!
    d_s^2\label{eq_theta-w_2},
\end{align}
where $\frac{\sinh \left(\left|w^{T}_n\right| / \rho\right)}{\cosh \left(\left|w^{T}_n\right| / \rho\right)} \in (-1,1)$, $w^{T}_n$ is the $n$-th element of $\bm w^T$ and $d_s$ denotes the number of non-zero elements in $\bm w$. Due to the $\mu_F$-strong convex of $F$, we have
\begin{align}
    &~~~~\mathbb{E}\left[\left\|\bm w^{T}-\bm w^{*}\right\|_2^{2}\right] \leq  \frac{2\left(\mathbb{E}\left[F\left({\bm w}_{T}\right)-F\left(\bm w^{*}\right)\right]\right)}{\mu_{F}}\label{eq_theta-w_3}.
\end{align}
By substituting \eqref{eq-t2-a}, \eqref{eq_theta-w_2} and \eqref{eq_theta-w_3} into \eqref{eq_theta-w} we complete the proof.
\end{proof}

\subsection*{Proof of Theorem~\ref{theorem_3}}
\begin{proof}

We first prove part (a). Similar with the proof in \cite{t2020personalized}, we rewrite the recursive formula due to the $L$-smooth of $F(\cdot)$
\begin{align}
&~\mathbb{E}\left[F\left(\bm w^{t+1}\right)-F\left(\bm w^{t}\right)\right] \notag\\
{\leq}&~ \mathbb{E}\left[\left\langle\nabla F\left(\bm w^{t}\right), \bm w^{t+1}-\bm w^{t}\right\rangle\right]+\frac{L_{F}}{2} \mathbb{E}\left[\left\|\bm w^{t+1}-\bm w^{t}\right\|_2^{2}\right] \notag\\
=&~-\tilde{\eta} \mathbb{E}\left[\left\langle\nabla F\left(\bm w^{t}\right), g^{t}\right\rangle\right]+\frac{\tilde{\eta}^{2} L_{F}}{2} \mathbb{E}\left[\left\|g^{t}\right\|_2^{2}\right] \notag\\
=&~-\tilde{\eta} \mathbb{E}\left[\left\|\nabla F\left(\bm w^{t}\right)\right\|_2^{2}\right]-\tilde{\eta} \mathbb{E}\left[\left\langle\nabla F\left(\bm w^{t}\right), g^{t}-\nabla F\left(\bm w^{t}\right)\right\rangle\right]\notag\\
&~+\frac{\tilde{\eta}^{2} L_{F}}{2} \mathbb{E}\left[\left\|g^{t}\right\|_2^{2}\right] \notag\\
\overset{(b)}{\leq}&~
-\tilde{\eta} \mathbb{E}\left[\left\|\nabla F\left(\bm w^{t}\right)\right\|_2^{2}\right]+
\frac{\tilde{\eta}}{2} \mathbb{E}\left[\left\|\nabla F\left(\bm w^{t}\right)\right\|_2^{2}\right]\notag\\
&~+\frac{\tilde{\eta}}{2} \mathbb{E}\left\|\frac{1}{N R} \sum_{i, r}^{N, R} g_{i, r}^{t}-\nabla F_{i}\left(\bm w^{t}\right)\right\|_2^{2}+\frac{\tilde{\eta}^{2} L_{F}}{2} \mathbb{E}\left[\left\|g^{t}\right\|_2^{2}\right] \notag\\
\overset{(c)}{\leq}&~
-\frac{\tilde\eta(1-3L_F\tilde\eta)}{2}\mathbb{E}\left[\|\nabla F(\bm w_t)\|_2^2\right] 
\notag\\
&~+\frac{\tilde\eta(1+3L_F\tilde\eta)}{2} \mathbb{E}\left\|\frac{1}{N R} \sum_{i, r}^{N, R} g_{i, r}^{t}-\nabla F_{i}\left(\bm w^{t}\right)\right\|_2^{2} \notag\\
&~+ \frac{3\tilde\eta^2 L_F}{2} \mathbb{E}\left\|\frac{1}{S} \sum_{i \in \mathcal{S}^{t}} \nabla F_{i}\left(\bm w^{t}\right)-\nabla F\left(\bm w^{t}\right)\right\|_2^{2}\notag\\
\notag\\
\overset{(d)}{\leq} &~
-\tilde{\eta} G\mathbb{E}\left[\left\|\nabla F\left(\bm w^{t}\right)\right\|_2^{2}\right] + \tilde{\eta}^{}G_1 + \tilde{\eta}^{2}G_2.\notag
\end{align}
where $(b)$ is due to the Cauchy-Swartz inequalities; $(c)$ is due to the Jensen's inequality; $(d)$ follows by $1+3L_F\tilde\eta\leq3\beta$ since $\tilde\eta\leq\frac{\beta}{2L_F}$, and $G_1 = 3\beta\bar\lambda^{2} \delta^{2}$, 
$G_2 = \frac{3L_F\left(49 R \sigma_{F, 2}^{2}+16 \delta^{2}\bar\lambda^{2}\right)}{R}$. 

Next,  let $\lambda^{2}-16 (L+\frac{\gamma_1}{\rho})^{2}\geq 1$ and $\tilde\eta \leq \frac{1}{588L_F\lambda^{2})}$ hold, we have
\begin{align}
    G&=\frac{1}{2}-\tilde{\eta} L_{F}\left(\frac{3}{2}+\frac{24 (L+\frac{\gamma_1}{\rho})^{2}}{\lambda^{2}-16 (L+\frac{\gamma_1}{\rho})^{2}} +\frac{72 \lambda^{2}}{\lambda^{2}-16 (L+\frac{\gamma_1}{\rho})^{2}}\right)\notag\\
    &\geq \frac{1}{2} - \frac{147\tilde{\eta} L_{F}\lambda^{2}}{2} \geq \frac{1}{4},\notag
\end{align}
we hence have
\begin{align}
	&~\frac{1}{T}\sum_{t=0}^{T-1}\mathbb{E}\left[\left\|\nabla F\left(\bm w^{t}\right)\right\|^{2}\right]\notag\\
	\leq&~ 4(\frac{\mathbb{E}\left[F\left(\bm w^{0}\right)-F\left(\bm w^{T}\right)\right]}{\tilde\eta T}+ G_1 + \tilde{\eta}^{1}G_2).\notag\\
	\leq&~ 4(\frac{\Delta_F}{\tilde\eta T}+ G_1 + \tilde{\eta}^{1}G_2),\notag
\end{align}
where $\Delta_F\triangleq F(\bm w^0)-F(W^*)$.

We next prove part (b), we have
\begin{align}
    &~\frac{1}{NJ} \sum_{i,j=1}^{NJ} \mathbb{E}\left[\left\|\tilde{\bm\theta}_{i,j}^{t}\left(\bm w^{t}\right)-\bm w^{t}\right\|_2^{2}\right]\notag\\
    \overset{(a)}{\leq}&~ \frac{4}{NJ} \sum_{i,j=1}^{NJ} \mathbb{E}\Bigg(\left\|\tilde{\theta}_{i,j}^{t}\!-\!\hat{\theta}_{i,j}^{t}\right\|_2^{2}\!+\!\left\|\hat{\theta}_{i,j}^{t}\!-\!\hat{\bm y}_{i,j}^{t} \!-\! \frac{\gamma_2}{\lambda_1}\nabla\phi_{\rho}(\boldsymbol{\bm w^t})\right\|_2^{2}\notag\\
    &~+\!\left\|\!\hat{\bm y}_{i,j}^{t} \!-\! \bm w^t \!-\! \frac{\gamma_2}{\lambda_2}\nabla\phi_{\rho}(\boldsymbol{\bm w^t})\right\|_2^{2} \!+\!\left\|\frac{\gamma_2(\lambda_1 \!+\! \lambda_2)}{\lambda_1\lambda_2}\nabla\phi_{\rho}(\boldsymbol{\bm w^t})\right\|_2^2\Bigg)\notag\\
    \overset{(b)}{\leq}&~ 4(\delta^{2}+\frac{\lambda_1^{2}+\lambda_2^{2}}{\lambda_1^{2}\lambda_2^{2}} \frac{1}{NJ} \sum_{i,j=1}^{NJ}\mathbb{E}\left[\left\|\nabla F_{i,j}\left(\bm w^{t}\right)\right\|_2^{2}\right]\notag\\
    &~+\frac{\gamma_2^2(\lambda_1+\lambda_2)^2}{\lambda_1^2\lambda_2^2}\frac{1}{NJ} \sum_{i,j=1}^{NJ}\mathbb{E}\left[\left\|\nabla\phi_{\rho}(\boldsymbol{\bm w^t})\right\|_2^2\right])\notag\\
    =&~ 4\left(\delta^{2}+\frac{1}{\lambda^2} \frac{1}{NJ} \sum_{i,j=1}^{NJ}\mathbb{E}\left[\left\|\nabla F_{i,j}\left(\bm w^{t}\right)\right\|_2^{2}\right]+\frac{\gamma_2^2}{\bar\lambda^2}d_s^2\right) \label{theorem3b-1}
\end{align}
where $(a)$ is due to the Jensen's inequality, $(b)$ is due to \eqref{first_order}. And the last term is bounded by
\begin{align}
    &~\frac{1}{NJ} \sum_{i,j=1}^{NJ}\mathbb{E}\left[\left\|\nabla F_{i,j}\left(\bm w^{t}\right)\right\|_2^{2}\right]\notag\\
    \overset{(a)}{=}&~\frac{1}{NJ} \sum_{i,j=1}^{NJ}\mathbb{E}\left[\left\|\nabla F_{i,j}\left(\bm w^{t}\right)-\nabla F_{}\left(\bm w^{t}\right)\right\|_2^{2}+\left\|\nabla F_{}\left(\bm w^{t}\right)\right\|_2^{2}\right]\notag\\
    \overset{(b)}{\leq}&~ \sigma_{F,2}^2+ \frac{\lambda^2}{\lambda^2-16(L+\frac{\gamma_1}{\rho})^2}\mathbb{E}\left[\left\|\nabla F_{}\left(\bm w^{t}\right)\right\|_2^{2}\right]\notag\\
    \overset{(c)}{\leq}&~ \sigma_{F,2}^2+\lambda^2\mathbb{E}\left[\left\|\nabla F_{}\left(\bm w^{t}\right)\right\|_2^{2}\right] \label{theorem3b-2}
\end{align}
where $(a)$ is due to the equation $\mathbb{E}\left[\|\bm x-\mathbb{E}\left[\bm x\right]\|_2^2\right]=\mathbb{E}\left[\|\bm x\|_2^2\right]-\mathbb{E}\left[\|\bm x\|_2\right]^2$, $(b)$ follows by Lemma~\ref{lemma_2}, $(c)$ is by setting $\lambda^2-16(L+\frac{\gamma_1}{\rho})^2\leq 1$.
By substituting \eqref{theorem3b-2} into \eqref{theorem3b-1}, we finished the proof.
\end{proof}


\bibliographystyle{IEEEtran}
\bibliography{ref.bib}

\begin{thebibliography}{10}
\providecommand{\url}[1]{#1}
\csname url@samestyle\endcsname
\providecommand{\newblock}{\relax}
\providecommand{\bibinfo}[2]{#2}
\providecommand{\BIBentrySTDinterwordspacing}{\spaceskip=0pt\relax}
\providecommand{\BIBentryALTinterwordstretchfactor}{4}
\providecommand{\BIBentryALTinterwordspacing}{\spaceskip=\fontdimen2\font plus
\BIBentryALTinterwordstretchfactor\fontdimen3\font minus \fontdimen4\font\relax}
\providecommand{\BIBforeignlanguage}[2]{{%
\expandafter\ifx\csname l@#1\endcsname\relax
\typeout{** WARNING: IEEEtran.bst: No hyphenation pattern has been}%
\typeout{** loaded for the language `#1'. Using the pattern for}%
\typeout{** the default language instead.}%
\else
\language=\csname l@#1\endcsname
\fi
#2}}
\providecommand{\BIBdecl}{\relax}
\BIBdecl

\bibitem{lecun2015deep}
Y.~LeCun, Y.~Bengio, and G.~Hinton, ``Deep learning,'' \emph{nature}, vol. 521, no. 7553, pp. 436--444, 2015.

\bibitem{chen2019deep}
J.~Chen and X.~Ran, ``Deep learning with edge computing: A review,'' \emph{Proc. of the IEEE}, vol. 107, no.~8, pp. 1655--1674, 2019.

\bibitem{mcmahan2017communication}
B.~McMahan, E.~Moore, D.~Ramage, S.~Hampson, and B.~A. y~Arcas, ``Communication-efficient learning of deep networks from decentralized data,'' in \emph{Proc. of Artificial Intelligence and Statistics}, 2017, pp. 1273--1282.

\bibitem{9233457}
W.~Zhang, Q.~Lu, Q.~Yu, Z.~Li, Y.~Liu, S.~K. Lo, S.~Chen, X.~Xu, and L.~Zhu, ``Blockchain-based federated learning for device failure detection in industrial iot,'' \emph{IEEE Internet of Things Journal}, vol.~8, no.~7, pp. 5926--5937, 2021.

\bibitem{9709603}
B.~Ghimire and D.~B. Rawat, ``Recent advances on federated learning for cybersecurity and cybersecurity for federated learning for internet of things,'' \emph{IEEE Internet of Things Journal}, vol.~9, no.~11, pp. 8229--8249, 2022.

\bibitem{9134426}
Y.~Shi, K.~Yang, T.~Jiang, J.~Zhang, and K.~B. Letaief, ``Communication-efficient edge ai: Algorithms and systems,'' \emph{IEEE Communications Surveys Tutorials}, vol.~22, no.~4, pp. 2167--2191, 2020.

\bibitem{9210812}
M.~Chen, Z.~Yang, W.~Saad, C.~Yin, H.~V. Poor, and S.~Cui, ``A joint learning and communications framework for federated learning over wireless networks,'' \emph{IEEE Transactions on Wireless Communications}, vol.~20, no.~1, pp. 269--283, 2021.

\bibitem{9477571}
J.~Le, X.~Lei, N.~Mu, H.~Zhang, K.~Zeng, and X.~Liao, ``Federated continuous learning with broad network architecture,'' \emph{IEEE Transactions on Cybernetics}, vol.~51, no.~8, pp. 3874--3888, 2021.

\bibitem{9079513}
S.~R. Pokhrel and J.~Choi, ``Federated learning with blockchain for autonomous vehicles: Analysis and design challenges,'' \emph{IEEE Transactions on Communications}, vol.~68, no.~8, pp. 4734--4746, 2020.

\bibitem{ferrag2021federated}
M.~A. Ferrag, O.~Friha, L.~Maglaras, H.~Janicke, and L.~Shu, ``Federated deep learning for cyber security in the internet of things: Concepts, applications, and experimental analysis,'' \emph{IEEE Access}, vol.~9, pp. 138\,509--138\,542, 2021.

\bibitem{khan2021federated}
L.~U. Khan, W.~Saad, Z.~Han, E.~Hossain, and C.~S. Hong, ``Federated learning for internet of things: Recent advances, taxonomy, and open challenges,'' \emph{IEEE Communications Surveys \& Tutorials}, vol.~23, no.~3, pp. 1759--1799, 2021.

\bibitem{wang2021federated}
Y.~Wang, G.~Gui, H.~Gacanin, B.~Adebisi, H.~Sari, and F.~Adachi, ``Federated learning for automatic modulation classification under class imbalance and varying noise condition,'' \emph{IEEE Transactions on Cognitive Communications and Networking}, vol.~8, no.~1, pp. 86--96, 2021.

\bibitem{he2021edge}
Z.~He, J.~Yin, Y.~Wang, G.~Gui, B.~Adebisi, T.~Ohtsuki, H.~Gacanin, and H.~Sari, ``Edge device identification based on federated learning and network traffic feature engineering,'' \emph{IEEE Transactions on Cognitive Communications and Networking}, vol.~8, no.~4, pp. 1898--1909, 2021.

\bibitem{li2019fedmd}
D.~Li and J.~Wang, ``Fedmd: Heterogenous federated learning via model distillation,'' \emph{arXiv preprint arXiv:1910.03581}, 2019.

\bibitem{deng2020adaptive}
Y.~Deng, M.~M. Kamani, and M.~Mahdavi, ``Adaptive personalized federated learning,'' \emph{arXiv preprint arXiv:2003.13461}, 2020.

\bibitem{sattler2021clustered}
F.~Sattler, K.-R. Müller, and W.~Samek, ``Clustered federated learning: Model-agnostic distributed multitask optimization under privacy constraints,'' \emph{IEEE Transactions on Neural Networks and Learning Systems}, vol.~32, no.~8, pp. 3710--3722, 2021.

\bibitem{arivazhagan2019federated}
M.~G. Arivazhagan, V.~Aggarwal, A.~K. Singh, and S.~Choudhary, ``Federated learning with personalization layers,'' \emph{arXiv preprint arXiv:1912.00818}, 2019.

\bibitem{li2018federated}
T.~Li, A.~K. Sahu, M.~Zaheer, M.~Sanjabi, A.~Talwalkar, and V.~Smith, ``Federated optimization in heterogeneous networks,'' \emph{arXiv preprint arXiv:1812.06127}, 2018.

\bibitem{t2020personalized}
C.~T.~Dinh, N.~Tran, and J.~Nguyen, ``Personalized federated learning with moreau envelopes,'' in \emph{Advances in Neural Information Processing Systems}, vol.~33, 2020, pp. 21\,394--21\,405.

\bibitem{huang2021personalized}
Y.~Huang, L.~Chu, Z.~Zhou, L.~Wang, J.~Liu, J.~Pei, and Y.~Zhang, ``Personalized cross-silo federated learning on non-iid data,'' in \emph{Proc. of Association for the Advancement of Artificial Intelligence}, vol.~35, no.~9, 2021, pp. 7865--7873.

\bibitem{fallah2020personalized}
A.~Fallah, A.~Mokhtari, and A.~Ozdaglar, ``Personalized federated learning: A meta-learning approach,'' \emph{arXiv preprint arXiv:2002.07948}, 2020.

\bibitem{achituve2021personalized}
I.~Achituve, A.~Shamsian, A.~Navon, G.~Chechik, and E.~Fetaya, ``Personalized federated learning with gaussian processes,'' in \emph{Advances in Neural Information Processing Systems}, vol.~34, 2021, pp. 8392--8406.

\bibitem{karimireddy2020scaffold}
S.~P. Karimireddy, S.~Kale, M.~Mohri, S.~Reddi, S.~Stich, and A.~T. Suresh, ``Scaffold: Stochastic controlled averaging for federated learning,'' in \emph{Proc. of International Conference on Machine Learning}, 2020, pp. 5132--5143.

\bibitem{liu2020client}
L.~Liu, J.~Zhang, S.~Song, and K.~B. Letaief, ``Client-edge-cloud hierarchical federated learning,'' in \emph{Proc. of IEEE International Conference on Communications}.\hskip 1em plus 0.5em minus 0.4em\relax IEEE, 2020, pp. 1--6.

\bibitem{bonawitz2019towards}
K.~Bonawitz, H.~Eichner, W.~Grieskamp, D.~Huba, A.~Ingerman, V.~Ivanov, C.~Kiddon, J.~Kone{\v{c}}n{\`y}, S.~Mazzocchi, H.~B. McMahan \emph{et~al.}, ``Towards federated learning at scale: System design,'' \emph{arXiv preprint arXiv:1902.01046}, 2019.

\bibitem{hsieh2020non}
K.~Hsieh, A.~Phanishayee, O.~Mutlu, and P.~Gibbons, ``The non-iid data quagmire of decentralized machine learning,'' in \emph{International Conference on Machine Learning}.\hskip 1em plus 0.5em minus 0.4em\relax PMLR, 2020, pp. 4387--4398.

\bibitem{kairouz2021advances}
P.~Kairouz, H.~B. McMahan, B.~Avent, A.~Bellet, M.~Bennis, A.~N. Bhagoji, K.~Bonawitz, Z.~Charles, G.~Cormode, R.~Cummings \emph{et~al.}, ``Advances and open problems in federated learning,'' \emph{Foundations and Trends{\textregistered} in Machine Learning}, vol.~14, no. 1--2, pp. 1--210, 2021.

\bibitem{mansour2020three}
Y.~Mansour, M.~Mohri, J.~Ro, and A.~T. Suresh, ``Three approaches for personalization with applications to federated learning,'' \emph{arXiv preprint arXiv:2002.10619}, 2020.

\bibitem{hinton2015distilling}
G.~Hinton, O.~Vinyals, and J.~Dean, ``Distilling the knowledge in a neural network,'' \emph{arXiv preprint arXiv:1503.02531}, 2015.

\bibitem{lin2020ensemble}
T.~Lin, L.~Kong, S.~U. Stich, and M.~Jaggi, ``Ensemble distillation for robust model fusion in federated learning,'' \emph{Advances in Neural Information Processing Systems}, vol.~33, pp. 2351--2363, 2020.

\bibitem{smith2017federated}
V.~Smith, C.-K. Chiang, M.~Sanjabi, and A.~S. Talwalkar, ``Federated multi-task learning,'' in \emph{Proc. of Advances in neural information processing systems}, 2017, pp. 4424--4434.

\bibitem{shoham2019overcoming}
N.~Shoham, T.~Avidor, A.~Keren, N.~Israel, D.~Benditkis, L.~Mor-Yosef, and I.~Zeitak, ``Overcoming forgetting in federated learning on non-iid data,'' \emph{arXiv preprint arXiv:1910.07796}, 2019.

\bibitem{7755794}
J.~{Sun}, Q.~{Qu}, and J.~{Wright}, ``Complete dictionary recovery over the sphere i: Overview and the geometric picture,'' \emph{IEEE Transactions on Information Theory}, vol.~63, no.~2, pp. 853--884, 2017.

\bibitem{beck2017first}
A.~Beck, \emph{First-order methods in optimization}.\hskip 1em plus 0.5em minus 0.4em\relax SIAM, 2017.

\bibitem{lecun1998gradient}
Y.~LeCun, L.~Bottou, Y.~Bengio, and P.~Haffner, ``Gradient-based learning applied to document recognition,'' \emph{Proc. of the IEEE}, vol.~86, no.~11, pp. 2278--2324, 1998.

\bibitem{xiao2017fashion}
H.~Xiao, K.~Rasul, and R.~Vollgraf, ``Fashion-mnist: a novel image dataset for benchmarking machine learning algorithms,'' \emph{arXiv preprint arXiv:1708.07747}, 2017.

\bibitem{krizhevsky2009learning}
A.~Krizhevsky, ``Learning multiple layers of features from tiny images,'' \emph{Master's thesis, University of Tront}, 2009.

\bibitem{liu2021personalized}
X.~Liu, Y.~Li, Q.~Wang, X.~Zhang, Y.~Shao, and Y.~Geng, ``Sparse personalized federated learning,'' \emph{IEEE Transactions on Neural Networks and Learning Systems}, pp. 1--15, 2023.

\end{thebibliography}

\vfill

\end{document}